\documentclass{article}

     \PassOptionsToPackage{numbers, compress}{natbib}
\usepackage[preprint]{neurips_2026}

\usepackage[utf8]{inputenc} 
\usepackage[T1]{fontenc}    
\usepackage{hyperref}       
\usepackage{url}            
\usepackage{booktabs}       
\usepackage{amsfonts}       
\usepackage{nicefrac}       
\usepackage{microtype}      
\usepackage{xcolor}         

\usepackage{amsmath}
\usepackage{multirow}
\usepackage{graphicx}
\usepackage{makecell}
\usepackage{colortbl}
\usepackage{adjustbox}
\usepackage{caption}
\usepackage{wrapfig}

\definecolor{color3}{rgb}{0.95,0.95,0.95}
\definecolor{color-ours}{HTML}{E8F7FF}

\title{DiffST: Spatiotemporal-Aware Diffusion for \\Real-World Space-Time Video Super-Resolution}

\author{
  Zheng Chen$^{1}$\thanks{Equal contribution.},\enspace 
  Ruofan Yang$^{1}$\footnotemark[1],\enspace 
  Jin Han$^{2}$,\enspace 
  Dehua Song$^{2}$,\enspace \\
  \textbf{Zichen Zou$^{1}$,}\enspace 
  \textbf{Chunming He$^{3}$,}\enspace 
  \textbf{Yong Guo$^{4}$,}\enspace 
  \textbf{Yulun Zhang$^{1}$}\thanks{Corresponding author: Yulun Zhang, yulun100@gmail.com} \\
  \textsuperscript{1}Shanghai Jiao Tong University,\enspace
  \textsuperscript{2}Huawei Noah's Ark Lab,\enspace\\
  \textsuperscript{3}Duke University,\enspace
  \textsuperscript{4}Huawei Consumer Business Group
  \vspace{-5.mm}
}

\begin{document}

\maketitle

\begin{abstract}
Diffusion-based models have shown strong performance in video super-resolution (VSR) and video frame interpolation (VFI). However, their role in the coupled space-time video super-resolution (STVSR) setting remains limited. Existing diffusion-based STVSR approaches suffer from two issues: \textbf{(1)} low inference efficiency and \textbf{(2)} insufficient utilization of spatiotemporal information. These limitations impede deployment. To address these issues, we introduce DiffST, an efficient spatiotemporal-aware video diffusion framework for real-world STVSR. To improve efficiency, we adapt a pre-trained diffusion model for one-step sampling and process the entire video directly rather than operating on individual frames. Furthermore, to enhance spatiotemporal information utilization, we introduce cross-frame context aggregation (CFCA) and video representation guidance (VRG). The CFCA module aggregates information across multiple keyframes to produce intermediate frames. The VRG module extracts video-level global features to guide the diffusion process. Extensive experiments show that DiffST obtains leading results on real-world STVSR tasks. It also maintains high inference efficiency, running about \textbf{17$\times$} faster than previous diffusion-based STVSR methods. Code is available at:~\url{https://github.com/zhengchen1999/DiffST}.
\end{abstract}

\vspace{-6.mm}
\section{Introduction}
\vspace{-3.mm}
Space-time video super-resolution (STVSR) aims to jointly increase spatial resolution and frame rate for spatially and temporally degraded videos~\cite{xu2021temporal,xiang2020zooming,chen2023motif,hu2023cycmunet+,kim2025bf}. Given the video compression that occurs during acquisition, transmission, and storage, as well as the limited quality of previously recorded video resources, STVSR has broad practical value. It can improve perceptual quality and motion smoothness, thereby improving the viewing experience in real applications.

One straightforward approach is to chain video super-resolution (VSR)~\cite{chan2021basicvsr,shi2022rethinking,chan2022investigating,zhou2024upscale,wang2025seedvr} and video frame interpolation (VFI)~\cite{huang2022real,danier2024ldmvfi,seo2025bim} methods. This pipeline can reach the target spatial and temporal resolutions. However, it overlooks the correlation between the two tasks, so spatiotemporal information is not fully shared across them, which limits the final restoration quality.

\begin{figure*}[t]
    \centering
    \includegraphics[width=\linewidth]{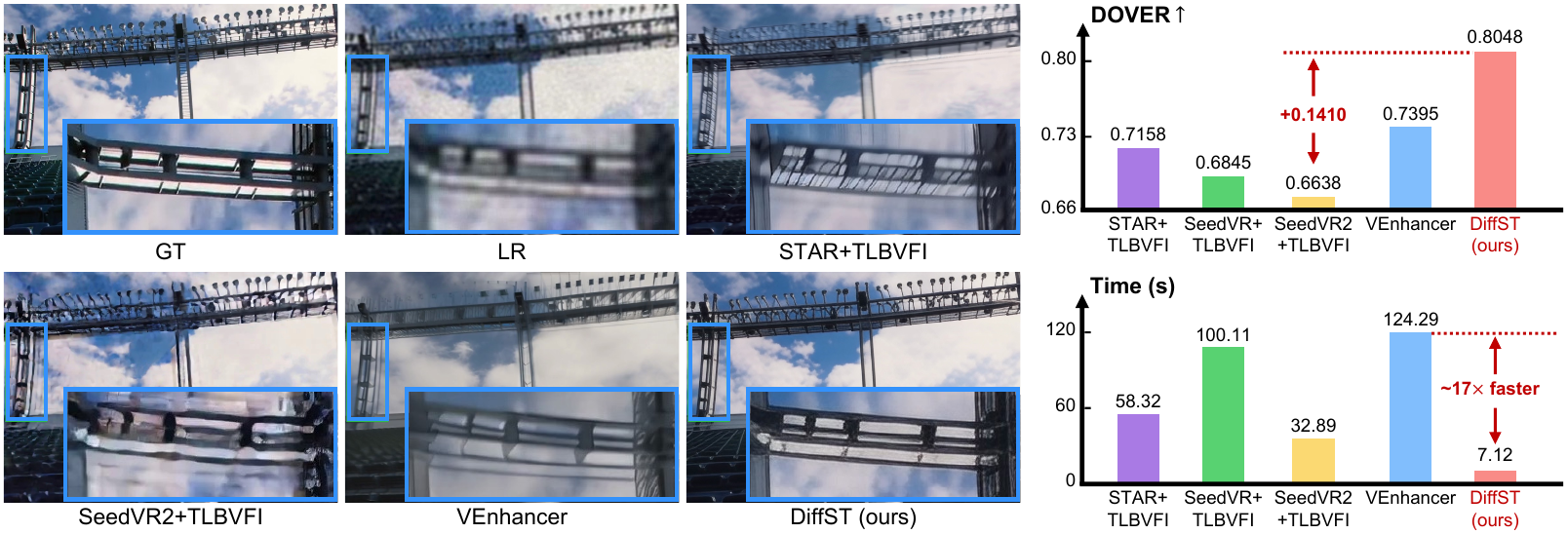}
    \vspace{-5.mm}
    \caption{Performance comparison of STVSR methods. The right-side quantitative scores are reported on the real-world dataset (RealVSR~\cite{yang2021real}). Runtime is measured on an A100 GPU using an output video of resolution 33$\times$720$\times$1280. Compared with VEnhancer~\cite{he2024venhancer}, DiffST achieves superior performance with an approximately \textbf{17$\times$} inference speedup.}
    \vspace{-5.mm}
    \label{fig:performance}
\end{figure*}

Therefore, researchers have developed unified frameworks that handle frame interpolation and super-resolution jointly within one stage~\cite{haris2020space,xu2021temporal,chen2023motif,kim2025bf,wei2025evenhancer}. These methods can better couple spatial and temporal cues. Nevertheless, many approaches follow the conventional VFI setting, \textit{i.e.}, processing only two keyframes (the first and last). Such a design has two drawbacks: 
\textbf{(1) Low efficiency.} Since each inference operates on two frames, the computational cost grows rapidly with video length.
\textbf{(2) Limited information.} Two-frame inputs cannot exploit the complementary information across multiple frames, which is crucial in complex motion or degradation scenarios.
Some studies attempt to overcome these issues by processing the entire video~\cite{xiang2020zooming,geng2022rstt}. 
However, restricted by model capabilities, the magnification is limited (\textit{e.g.}, $\times$2), and severe detail loss occurs.

Recently, large pre-trained video diffusion models have shown impressive generative capability~\cite{yang2024cogvideox,kong2024hunyuanvideo,wan2025wan}. They have also been introduced into VSR~\cite{xie2025star,zhuang2025flashvsr} and VFI~\cite{seo2025bim,wang2024generative,yang2024vibidsampler} tasks, demonstrating great potential. Compared with previous methods (\textit{e.g.}, convolution-based models), they can produce more realistic details. For STVSR, some methods leverage pre-trained models by treating the low-quality video as a condition~\cite{he2024venhancer}. This paradigm enables whole-video processing, mitigating issues present in previous methods. However, two major limitations remain: 
\textbf{(1) Low efficiency.} The model takes random noise as input and conditions on the video. As a result, it requires multiple sampling steps to produce clear results, leading to slow inference.
\textbf{(2) Limited information.} Conditioning uses video spatiotemporal cues only implicitly, which prevents full exploitation of the video information.
These limitations hinder the application of STVSR methods in the real world.

To address the above issues, we propose DiffST, an efficient spatiotemporal-aware video diffusion model for real-world STVSR. DiffST is based on a pre-trained video generation model (\textit{i.e.}, WAN~\cite{wan2025wan}) to exploit its rich generative prior. Meanwhile, it is specifically designed to overcome the inefficiency and limited information utilization of existing approaches.
First, to improve efficiency, we process the entire video directly (instead of frame by frame) and adopt it as the model input (instead of the condition). This design enables the model to leverage the rich structural information contained in the input video. Thus, we can compress the multi-step sampling process into a single step, greatly enhancing inference efficiency. It also eliminates the need for heavy conditional modules, such as ControlNet~\cite{zhang2023adding,he2024venhancer}, while preserving the generative prior.

Moreover, to enhance information utilization, we design two modules: cross-frame context aggregation and video representation guidance. These modules explicitly exploit spatiotemporal information to enhance video restoration. 
\textbf{(1) Aggregation.} To better leverage temporal information, we fuse multiple keyframes to generate an intermediate frame. The multi-to-one manner allows the model to leverage broader contextual cues and better handle challenging cases such as severe degradation.
\textbf{(2) Guidance.} We further extract representations from multiple keyframes and fuse them into a global video representation. This video representation then guides the diffusion generation process, providing explicit spatiotemporal cues throughout restoration.

Benefiting from our one-step, video-level inference paradigm, and the proposed aggregation and guidance modules, DiffST achieves outstanding restoration performance and efficiency. As shown in Fig.~\ref{fig:performance}, compared with recent diffusion-based STVSR and VSR+VFI approaches, our method exhibits significant advantages in spatial detail, temporal consistency, and inference speed. Compared with VEnhancer~\cite{he2024venhancer}, it attains a \textbf{17$\times$} speedup under the same setting.

Overall, our contributions can be summarized as follows:
\vspace{-1.mm}
\begin{itemize}
\item We propose DiffST, a diffusion-based space-time video super-resolution model for real-world scenarios. The one-step, video-level inference ensures high efficiency.

\item We introduce cross-frame context aggregation and video representation guidance to leverage spatiotemporal information, improving detail and consistency.

\item Extensive experiments on synthetic and real-world STVSR datasets demonstrate the superior restoration performance and efficiency of our method.
\end{itemize}

\section{Related Work}
\subsection{Video Super-Resolution}
Previous video super-resolution (VSR) approaches~\cite{chan2021basicvsr,chan2022basicvsr++,kappeler2016video,liao2015video,liu2017robust} typically use optical flow to exploit temporal information, but suffer from flow estimation errors and the complexity of two-stage models. Several studies~\cite{tian2020tdan,jo2018deep} opt to forgo optical flow, adopting end-to-end models. For instance, VSR-DUF~\cite{jo2018deep} learns dynamic filters and residuals directly from the input. Recently, with the rapid progress of generative models, diffusion-based methods~\cite{li2025diffvsr,yang2024motion,wang2025seedvr,zhou2024upscale,zhang2025infvsr} have become an increasingly adopted direction for VSR. 
UAV~\cite{zhou2024upscale} uses a pretrained image diffusion model and inserts temporal layers into the diffusion.
STAR~\cite{xie2025star} augments a text-to-video diffusion model with an enhancement module to capture spatial details. Going further, one-step diffusion methods~\cite{chen2025dove,wang2025seedvr2,li2025asymmetric} seek to alleviate the burden of multi-step diffusion denoising. For instance, SeedVR2~\cite{wang2025seedvr2} proposes an efficient one-step diffusion transformer that employs the window attention. 

\subsection{Video Frame Interpolation}
Conventional learning-based video frame interpolation (VFI) methods include optical flow estimation and kernel prediction. Flow-based methods~\cite{niklaus2018context,huang2022real,jin2023unified} model motion by estimating optical flow, whereas kernel-based approaches~\cite{zhou2023exploring,liu2017video,lee2020adacof,ding2021cdfi} learn per-pixel adaptive kernels. Recently, diffusion models are increasingly used for VFI~\cite{lyu2025tlb,lew2025disentangled,danier2024ldmvfi}. MoMo~\cite{lew2025disentangled} generates intermediate bi-directional optical flows via a dedicated motion diffusion model. LDMVFI~\cite{danier2024ldmvfi} uses a VFI-specific autoencoder and a denoising U-Net to generate intermediate frames. 
EDEN~\cite{zhang2025eden} replaces the U-Net structure with the diffusion transformer to avoid information loss. 
However, many VFI models are limited to two-frame input, making it difficult to exploit long-range temporal cues from the video. This limitation also reduces their flexibility in real-world.

\subsection{Space-Time Video Super Resolution}
Space-time video super-resolution (STVSR) refers to reconstructing spatially detailed, high-frame-rate videos from sparsely sampled input. Early methods~\cite{shechtman2002increasing,mudenagudi2010space} model low-resolution videos as degraded observations of HR scenes corrupted by blur, sampling, and motion. By contrast, CNN-based methods~\cite{haris2020space,xu2021temporal,cao2022towards} view STVSR as an end-to-end learning problem. 
 STARnet~\cite{haris2020space} integrates multi-resolution spatial–temporal features for detail enhancement and motion estimation. 
Zooming Slow-Mo~\cite{xiang2020zooming} employs deformable feature interpolation and a deformable ConvLSTM to jointly perform spatial and temporal upscaling. 
In recent work, to leverage the powerful generative capabilities of diffusion models, VEnhancer~\cite{he2024venhancer} builds on video diffusion models with a trainable video ControlNet. Nevertheless, insufficient use of spatiotemporal cues restricts STVSR performance, particularly in two-frame settings~\cite{chen2022videoinr,haris2020space,chen2023motif}. Moreover, diffusion-based STVSR~\cite{he2024venhancer} suffers from low inference efficiency due to multi-step sampling at inference time.

\begin{figure*}[t]
    \centering
    \includegraphics[width=\linewidth]{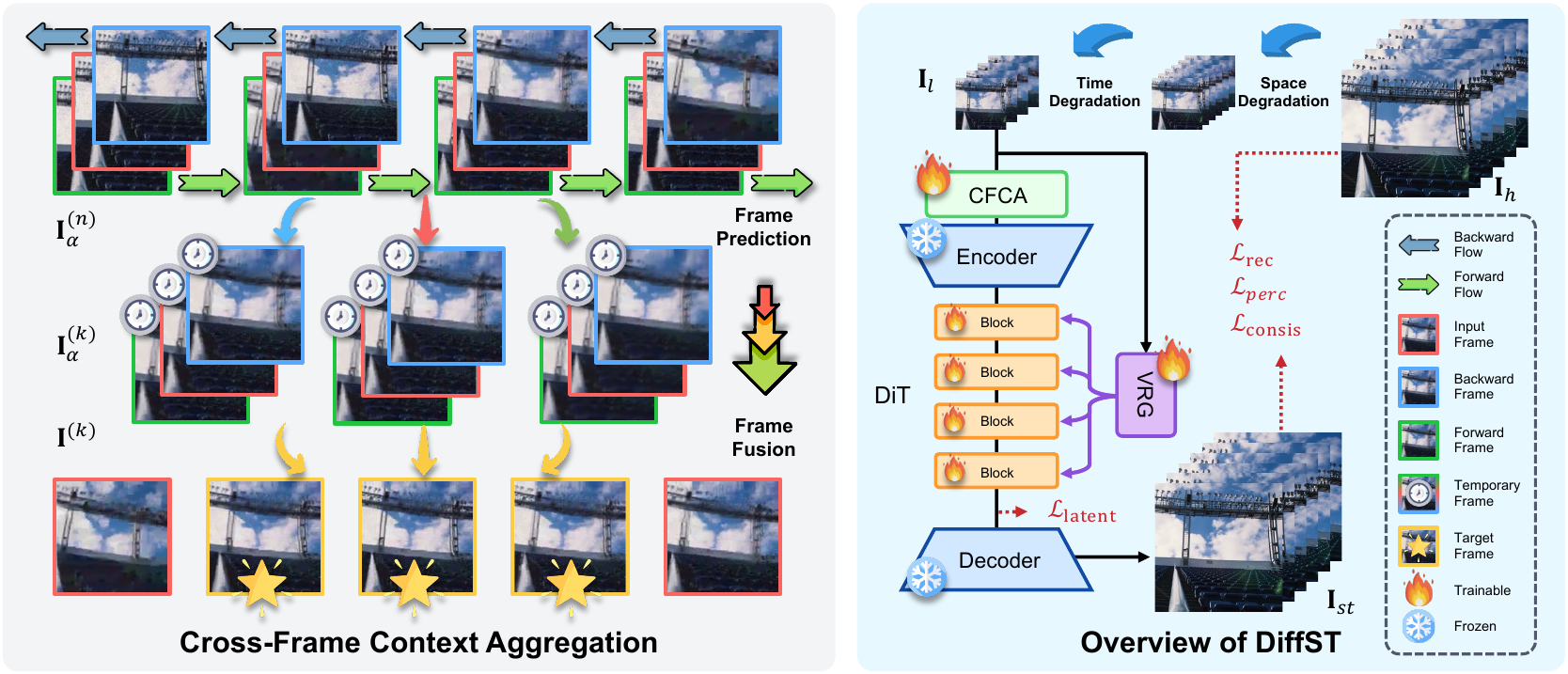}
    \caption{Overview of the proposed DiffST. Built upon a pre-trained video generation diffusion model, DiffST performs one-step sampling to process the entire video directly. The cross-frame context aggregation (CFCA) module aggregates information from multiple frames to generate intermediate frames. The video representation guidance (VRG) module extracts a video-level representation to guide the restoration process. To better match real-world conditions, we adopt multiple spatial degradations, combined with frame subsampling (temporal).}
    \vspace{-4.mm}
    \label{fig:pipeline}
\end{figure*}

\vspace{-1.mm}
\section{Method}
\vspace{-1.mm}
In this section, we present DiffST, our diffusion-based approach for space-time video super-resolution (STVSR). We first define the real-world video-level STVSR problem. Then, we describe the overall model architecture. Finally, we introduce the proposed cross-frame context aggregation (CFCA) and video representation guidance (VRG) modules used in DiffST.

\subsection{Problem Setting}
\label{sec:problem}
The goal of space-time video super-resolution (STVSR) is to enhance low spatial and temporal resolution video, improving spatial detail and motion smoothness. 
A common problem setting takes the first and last frames of a short video clip (\textit{e.g.}, 7 frames), downsamples them through interpolation to serve as inputs. The STVSR model reconstructs the video from these two low-resolution frames.

This setting intuitively combines the \textbf{classic} video super-resolution (VSR) and video frame interpolation (VFI) tasks. 
However, this task definition differs from that in the real world. In practice, the video data to be processed contains multiple frames rather than only two keyframes. Moreover, during capture and transmission, videos often suffer from various degradations (\textit{e.g.}, blur or noise). Simple interpolation downsampling is insufficient to represent reality.

Thus, we define a new video-level STVSR task that better aligns with real-world scenarios. Given an input video with low spatial resolution and low frame rate $\mathbf{I}_{l}$$\in$$\mathbb{R}^{\frac{T}{\varphi_t} \times \frac{H}{\varphi_s} \times \frac{W}{\varphi_s} \times 3}$, the objective is to recover a high-resolution and high-frame-rate output $\mathbf{I}_{st}$$\in$$\mathbb{R}^{T \times H \times W \times 3}$, where $T$ is the frame number, $H$$\times$$W$ denotes the spatial resolution, and $\varphi_s$ and $\varphi_t$ are spatial and temporal scaling factors:
\begin{equation}
\begin{gathered}
\mathbf{I}_{st} = \mathcal{F}_{\theta}(\mathbf{I}_l; \varphi_s, \varphi_t),
\end{gathered}
\end{equation}
where, $\mathcal{F}$ denotes the STVSR model. To simulate realistic degradations, we construct $\mathbf{I}_{l}$ from the high-quality ground truth video $\mathbf{I}_{h}$ with the degradation process illustrated in Fig.~\ref{fig:pipeline}. Specifically, $\mathbf{I}_{h}$ undergoes multiple degradation operations to generate the low-resolution sequence. Then, frames are temporally sampled with a sliding window according to the temporal sacle $\varphi_t$, resulting $\mathbf{I}_{l}$. This formulation preserves the input as a continuous video clip rather than isolated endpoints, allowing the model to exploit broader temporal context during restoration.

\vspace{-2.mm}
\subsection{Model Overview}
\vspace{-1.mm}
To solve the STVSR task, we propose DiffST, as illustrated in Fig.~\ref{fig:pipeline}. Our model builds on the pre-trained video generation model (\textit{i.e.}, WAN~\cite{wan2025wan}).
Given the low-resolution and low-frame-rate video $\mathbf{I}_{l}$, we first employ the cross-frame context aggregation module to exploit information from multiple frames and predict the target intermediate frames. Then, we apply bilinear interpolation to upsample the results by a factor ($\varphi_s$), obtaining the intermediate $\mathbf{I}_{m}$ that matches the target ($\mathbf{I}_{h}$).

The intermediate video $\mathbf{I}_{m}$ is compressed into a latent representation $\mathbf{z}$ through the VAE encoder. Then, the target refined latent $\mathbf{z}_{st}$ is produced through single-step sampling by the transformer-based velocity prediction network $\epsilon_{\theta}$. Since the pre-trained model follows the flow-matching diffusion architecture with the Euler ODE solver, the sampling process is expressed as:
\begin{equation}
\begin{gathered}
\mathbf{z}_{st} = \mathbf{z} - \sigma_{t} \mathcal{V}_{\theta}(\mathbf{z}, t, \mathbf{c}),
\label{eq:zst}
\end{gathered}
\end{equation}
where $t$ is the timestep, $\sigma_{t}$ denotes the noise level, and $\mathbf{c}$ represents the condition prompt. By initializing sampling from the structure-rich latent $\mathbf{z}$ (instead of random noise), single-step sampling can yield target reconstructions without iterative denoising from pure noise.

Meanwhile, to further enhance restoration quality, we introduce the video representation guidance module to provide global video-aware cues. The module extracts the global video representation from the input $\mathbf{I}_{l}$ as the prompt $\mathbf{c}$. This explicit utilization of video-level information constrains the diffusion sampling. Unlike frame-wise prompts, this video-level representation summarizes the overall temporal context and provides consistent guidance for the entire sequence.
Finally, the refined latent $\mathbf{z}_{st}$ is decoded through the VAE decoder to obtain the final STVSR output $\mathbf{I}_{st}$. Next, we describe the two explicit spatiotemporal information utilization modules in detail.

\begin{wrapfigure}{r}{0.53\textwidth}
    \centering
    \vspace{-11.mm}
    \includegraphics[width=\linewidth]{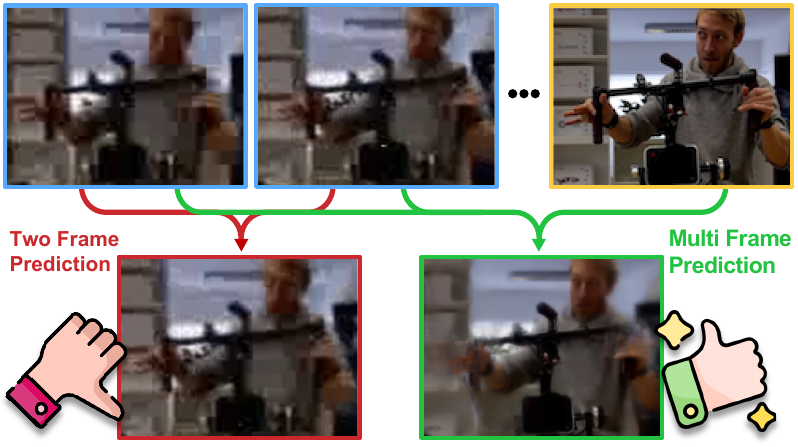}
    \vspace{-5.mm}
    \caption{The visualization shows that leveraging multiple frames to predict intermediate frames produces clearer results in the complex scenarios. In contrast, applying only two frames lacks temporal information and leads to inferior outcomes.}
    \vspace{-5.mm}
    \label{fig:aggregation}
\end{wrapfigure}

\subsection{Cross-Frame Context Aggregation}
To increase frame rate, video frame interpolation (VFI) methods typically predict intermediate frames from two adjacent keyframes, either by direct regression or by flow-based warping. Compared with linear interpolation, these approaches achieve more accurate predictions under mild motion and clean inputs. 

However, under complex motion or severe degradation, relying only on adjacent frames remains insufficient. As shown in Fig.~\ref{fig:aggregation}, in the first two frames, the person appears already heavily degraded with substantial detail loss. The intermediate frames predicted from these two frames also exhibit severe detail loss. In contrast, the content of subsequent frames is relatively clear. This observation motivates us to aggregate information from multiple frames to fully utilize temporal cues for more accurate predictions.

Based on this insight, we propose the cross-frame context aggregation module, as displayed in Fig.~\ref{fig:pipeline}. We first propagate information across frames by flow-based warping. For the input video $\mathbf{I}_{l}$, we estimate both forward and backward flow. Using these flows, we perform temporal propagation by fusing neighboring frames. Specifically, given the forward flow, the backward fused video $\mathbf{I}_b$ is:
\begin{equation}
\mathbf{I}_b^{(n)} = \mathcal{W}\!\big(\mathbf{I}_b^{(n+1)}, \mathbf{F}^{n\rightarrow n+1}\big)\odot \mathbf{M}_n  + \mathbf{I}_{l}^{(n)} \odot \big(1 - \mathbf{M}_n\big),
\label{eq:propagat}
\end{equation}
where $n$ denotes the frame index; $\mathbf{F}_{l}^{n\rightarrow n+1}$ means the forward flow from input video $\mathbf{I}_{l}$; $\mathcal{W}(\cdot,\cdot)$ represents the backward warping operation, and $\mathbf{M}$ is a validity mask derived from forward-backward flow consistency~\cite{xu2019deep,zhou2023propainter}. This fusion strategy can effectively propagate information and suppress unreliable regions. By reversing Eq.~\eqref{eq:propagat}, we can obtain the forward fused video $\mathbf{I}_f$.

Based on the fused videos $\mathbf{I}_f$ and $\mathbf{I}_b$, and the input video $\mathbf{I}_l$, we predict the intermediate frames from complementary temporal contexts. For the $k$-th intermediate frame $\mathbf{I}^{(k)}$, let its neighboring keyframes be $m$ and $m$$+$$1$. The prediction process is defined as:
\begin{equation}
\begin{aligned}
\mathbf{I}_{\alpha}^{(k)} 
&= (1-\mathbf{S}_{\alpha}^{(k)}) \odot 
   \mathcal{W}\!\big(\mathbf{I}_{\alpha}^{(m+1)}, \mathbf{F}_{\alpha}^{k\to m+1}\big) \\
&\quad + \mathbf{S}_{\alpha}^{(k)} \odot 
   \mathcal{W}\!\big(\mathbf{I}_{\alpha}^{(m)}, \mathbf{F}_{\alpha}^{k\to m}\big),
   \quad \alpha \!\in\! \{l,f,b\},
\end{aligned}
\end{equation}
where $\mathbf{I}_{\alpha}^{(k)}$ are the temporary frames predicted from three videos, $\mathbf{F}_{\alpha}$ is the corresponding optical flow, and $\mathbf{S}_{\alpha}^{(k)}$ is the fusion map. The $\mathbf{S}_{\alpha}^{(k)}$, $\mathbf{F}_{\alpha}^{k\to m+1}$, and $\mathbf{F}_{\alpha}^{k\to m}$ are estimated from keyframes through the network~\cite{huang2022real}. 
Finally, we fuse the three frames to obtain the final target $\mathbf{I}^{(k)}$:
\begin{equation}
\mathbf{I}^{(k)} = \Psi_{\theta}\big(\mathbf{I}_{l}^{(k)}, \mathbf{I}_{f}^{(k)}, \mathbf{I}_{b}^{(k)}\big),
\end{equation}
where $\Psi_{\theta}$ is a learnable fusion network. By integrating rich information from multiple frames, our aggregation module produces more accurate intermediate frames, providing reliable inputs for subsequent processing by the diffusion backbone.

\vspace{-2.mm}
\subsection{Video Representation Guidance}
\vspace{-2.mm}
The prompt plays a crucial role in the diffusion generation process. Thus, for video restoration, some studies retain the original T2V model condition, using text prompts with a fixed string~\cite{chen2025dove,zhuang2025flashvsr} or extracted from the first frame~\cite{zhou2024upscale} via a vision-language model (VLM). This matches the pre-training distribution and stabilizes training. However, it ignores temporal information and cannot represent the whole video, which limits the guidance. Besides, some methods adopt ControlNet~\cite{xie2025star,zhang2023adding} to provide conditions. Nevertheless, the module approaches introduce significant overhead.

\begin{wrapfigure}{r}{0.60\textwidth}
    \centering
    \vspace{-11.mm}
    \includegraphics[width=\linewidth]{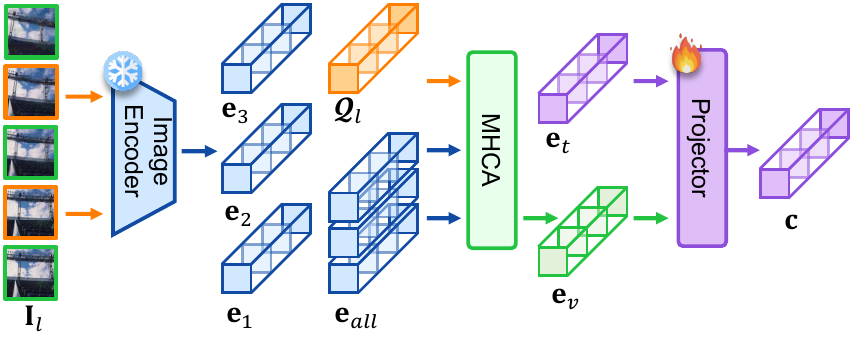}
    \caption{We select keyframes from the input video, encode them, and fuse the features to construct a video representation. This representation is combined with text embeddings to align with the pre-trained diffusion distribution.}
    \vspace{-4.mm}
    \label{fig:guidance}
\end{wrapfigure}

To overcome these limitations, we propose the video representation guidance module $\Phi_{\theta}$, as illustrated in Fig.~\ref{fig:guidance}. The module extracts global video-level prompt embedding from the input $\mathbf{I}_l$ to guide the diffusion process. We feed the embedding through the built-in prompt guidance path (\textit{i.e.}, cross-attention) to maintain efficiency without extra conditional networks.

Specifically, we select $N_k$ keyframes uniformly from the input video $\mathbf{I}_l$ to cover the temporal span. Each keyframe is encoded through a pre-trained image encoder, \textit{i.e.}, DAPE~\cite{wu2023seesr} (denoted as $\mathcal{E}_{\theta}$), to obtain spatial representations from representative frames. To combine spatial cues with temporal context, we introduce the multi-head attention module to aggregate these $N_k$ frame embeddings into a unified video representation $\mathbf{e}_v$. The process is formulated as:
\begin{equation}
\begin{gathered}
\{\mathbf{I}_l^{(k)}\}_{k=1}^{N} = \mathcal{S}(\mathbf{I}_l, N), \quad
\mathbf{e}_k = \mathcal{E}_{\theta}(\mathbf{I}_l^{(k)}), \\
\mathbf{e}_{all} = [\mathbf{e}_1, \ldots, \mathbf{e}_N], \quad
\mathbf{e}_v = \mathrm{MHCA}(\mathbf{Q}_l, \mathbf{e}_{all}, \mathbf{e}_{all}),
\label{eq:video_prompt}
\end{gathered}
\end{equation}
where $\mathcal{S}(\cdot, N)$ denotes uniform sampling, $\mathrm{MHCA}(Q,K,V)$ means the multi-head cross-attention operation, and $\mathbf{Q}_{l}$ is a learnable parameter. Moreover, since the video representation $\mathbf{e}_v$ differs from the original diffusion embedding space, we further fuse $\mathbf{e}_v$ with the text embedding $\mathbf{e}_t$ to generate the final prompt embedding. The text embedding $\mathbf{e}_t$ comes from a fixed description for efficiency. This reduces training difficulty and provides additional semantic cues while keeping the prompt pathway lightweight. Therefore, the final condition prompt $\mathbf{c}$ can be calculated as:
\begin{equation}
\mathbf{c} = \mathcal{P}_{\theta}\big([\mathbf{e}_v, \mathbf{e}_t]\big),
\label{eq:video_text_prompt}
\end{equation}
where $\mathcal{P}_{\theta}$ is a learnable projector to fuse the embedding. The resulting video prompt embedding $\mathbf{c}$ is then applied in Eq.~\eqref{eq:zst} to guide diffusion. It offers explicit global spatiotemporal guidance, improving restoration results across the multi temporal sequence.

\subsection{Training Objectives}
We employ a set of loss functions to optimize the velocity prediction network $\mathcal{V}_{\theta}$, the intermediate frame fusion network $\Psi_{\theta}$, and the video representation extraction module $\Phi_{\theta}$. All other components, such as the VAE, are kept frozen for adaptation.

In the \textbf{latent domain}, we compute the MSE loss between the latent $\mathbf{z}_{h}$ from the ground-truth $\mathbf{I}_{h}$ and the predicted latent $\mathbf{z}_{st}$ obtained from Eq.~\eqref{eq:zst} to supervise the one-step refinement and align the restored latent with the target video distribution. The loss is computed as:
\begin{equation}
\mathcal{L}_{\text{latent}} = \mathcal{L}_{\text{MSE}}(\mathbf{z}_{st}, \mathbf{z}_{h}) = \frac{1}{\mathbf{z}_{st}}||\mathbf{z}_{st} - \mathbf{z}_{h}||_2^2.
\end{equation}

In the \textbf{pixel domain}, we calculate the reconstruction loss (MSE) and a perceptual loss (LPIPS~\cite{zhang2018unreasonable}) between the DiffST output $\mathbf{I}_{st}$ and the ground-truth $\mathbf{I}_{h}$:
\begin{equation}
\mathcal{L}_{\text{rec}} = \mathcal{L}_{\text{MSE}}(\mathbf{I}_{st}, \mathbf{I}_{h}), \quad \mathcal{L}_{\text{perc}} = \mathcal{L}_{\text{LPIPS}}(\mathbf{I}_{st}, \mathbf{I}_{h}).
\end{equation}

To further enhance temporal coherence, we introduce a bidirectional temporal consistency loss. We extract forward and backward optical flows $\mathbf{F}_{h}$ from ground-truth video $\mathbf{I}_{h}$, and compute the loss consistency between the warped frame and its corresponding frame:
\begin{equation}
\begin{aligned}
\mathcal{L}_{\text{consis}} = \sum_i \big(
& \|\mathcal{W}(\mathbf{I}_{st}^{(i+1)}, \mathbf{F}_{h}^{i+1\to i}) - \mathbf{I}_{st}^{i}\|_1 \\
&+ \|\mathcal{W}(\mathbf{I}_{st}^{(i-1)}, \mathbf{F}_{h}^{i-1\to i}) - \mathbf{I}_{st}^{i}\|_1 \big).
\end{aligned}
\end{equation}
This loss encourages consistency between adjacent frames and suppresses temporal flickering in the restored video. Finally, our overall loss is a weighted sum:
\begin{equation}
\mathcal{L}_{\text{final}} = \mathcal{L}_{\text{latent}} + \mathcal{L}_{\text{rec}} + \mathcal{L}_{\text{perc}} + \gamma_{consis} \mathcal{L}_{\text{consis}},
\label{eq:loss}
\end{equation}
where $\mathcal{L}_{\text{consis}}$ is the loss scaler. With this joint loss formulation, we train DiffST end-to-end to realize efficient, video-level, real-world space-time video super-resolution.

\vspace{-1.mm}
\section{Experiments}
\label{sec:experiments}
\vspace{-2.mm}
\subsection{Experimental Settings}
\vspace{-2.mm}
\label{sec:setting}
\noindent \textbf{Datasets.}
We apply HQ-VSR~\cite{chen2025dove} as the training dataset. The training LQ-HQ pairs are generated following the degradation pipeline in Sec.~\ref{sec:problem}, where the spatial degradation follows RealBasicVSR~\cite{chan2022investigating}. Both spatial and temporal scaling factors ($\varphi_s$, $\varphi_t$) are set to 4. Evaluation is conducted on synthetic (UDM10~\cite{tao2017detail} and Vid4~\cite{liu2011bayesian}) and real-world (MVSR4x~\cite{wang2023benchmark} and RealVSR~\cite{yang2021real}) benchmarks. For synthetic datasets, we reuse the training degradation; for real-world benchmarks, only temporal subsampling is applied for fair comparison across methods.

\noindent \textbf{Evaluation Metrics.}
We employ multiple metrics to assess fidelity, perceptual quality, and temporal consistency. For fidelity, PSNR and SSIM~\cite{wang2004image} are reported. For perceptual quality, we evaluate with LPIPS~\cite{zhang2018unreasonable}, DISTS~\cite{ding2020image}, CLIP-IQA~\cite{wang2023exploring}, MUSIQA~\cite{ke2021musiq}, and MANIQA~\cite{yang2022maniqa}. For general video quality and temporal consistency, DOVER~\cite{wu2023exploring} is adopted. Additional temporal-consistency and perceptual video metrics are defined in Appendix~\ref{sec:app_more_metric_def}, and their scores are given in Appendix~\ref{sec:app_more_metric_results}.

\noindent \textbf{Implementation Details.}
The backbone is the pre-trained Wan2.1-T2V-1.3B~\cite{wan2025wan}. In one-step inference, we set $t$$=$$799$. The video representation guidance module takes $N_k$$=$$5$ keyframes.
We adopt the loss in Eq.~\eqref{eq:loss}, where $\mathcal{L}_{\text{consis}}$ is $0.1$. AdamW is used ($\beta_1$$=$$0.9$, $\beta_2$$=$$0.999$) with a learning rate of 5$\times$10$^{-5}$. Training videos are cropped to 17$\times$320$\times$640. The batch size is 4 and the iteration is 10,000. Experiments are conducted on four A100 GPU.

\begin{table}[t]
\scriptsize
\centering
\begin{minipage}[t]{0.49\linewidth}
\centering
\setlength{\tabcolsep}{1.5mm}
\begin{tabular}{l|cccc} 
        \toprule
        \rowcolor{color3} Method & PSNR $\uparrow$ & LPIPS $\downarrow$ & CLIP-IQA $\uparrow$ & DOVER $\uparrow$\\
        \midrule
        Baseline & 24.49 & 0.2757 & 0.3859 & 0.7081 \\
        $+$Aggregation & 24.87 & 0.2609 & 0.4002 & 0.7564 \\
        $+$Guidance & \textbf{24.92} & \textbf{0.2564} & \textbf{0.4086} & \textbf{0.7780} \\
        \bottomrule
\end{tabular}
\vspace{1.mm}
\caption{Break down ablation.}
\label{tab:abl_break}
\end{minipage}
\hfill
\begin{minipage}[t]{0.49\linewidth}
\centering
\setlength{\tabcolsep}{1.5mm}
\begin{tabular}{l|cccc} 
        \toprule
        \rowcolor{color3} Aggregation & PSNR $\uparrow$ & LPIPS $\downarrow$ & CLIP-IQA $\uparrow$ & DOVER $\uparrow$ \\
        \midrule
        Interpolation & 24.49 & 0.2757 & 0.3859 & 0.7081 \\
        Flow (Two)  & 24.75 & 0.2637 & 0.3910 & 0.7342  \\
        Flow (Multi) & \textbf{24.87} & \textbf{0.2609} & \textbf{0.4002} & \textbf{0.7564} \\
        \bottomrule
\end{tabular}
\vspace{1.mm}
\caption{Ablation on aggregation method.}
\label{tab:abl_aggregation}
\end{minipage}
\vspace{-8.mm}
\end{table}

\vspace{-2.5mm}
\subsection{Ablation Study}
\vspace{-2.5mm}
We evaluate the effectiveness of our method through ablations. All training settings follow Sec.~\ref{sec:setting}. We evaluate on UDM10~\cite{tao2017detail} for controlled comparison of core components. 

\noindent \textbf{Break Down.}
We perform a breakdown ablation to examine the contribution of each component. The results are shown in Tab.~\ref{tab:abl_break}. The baseline adopts the original WAN model (\textit{i.e.}, Wan2.1-T2V-1.3B)~\cite{wan2025wan}, where inputs are upsampled to the target size via interpolation. Gradually incorporating the proposed cross-frame context aggregation (\textit{i.e.}, Aggregation) and video representation guidance (\textit{i.e.}, Guidance) modules brings consistent improvements across multiple dimensions. Compared with the baseline, the complete model improves PSNR by 0.43 dB.

\noindent \textbf{Cross-Frame Context Aggregation.}
We perform an ablation on the aggregation module, with results shown in Tab.~\ref{tab:abl_aggregation}. Interpolation denotes generating intermediate frames via linear interpolation. Flow (Two) predicts intermediate frames using optical flow~\cite{huang2022real} between two adjacent frames. Flow (Multi) corresponds to our proposed cross-frame context aggregation. The results show that our multi-frame aggregation improves DOVER by 0.0222 over Flow (Two) and 0.0483 over Interpolation. This demonstrates that our proposed multi-frame aggregation is able to leverages spatiotemporal information, producing clearer intermediate results that better support subsequent processing.

\begin{wraptable}{r}{0.5\textwidth}
\scriptsize
\centering
\vspace{-4.mm}
\setlength{\tabcolsep}{1.7mm}
\begin{tabular}{l|cccc} 
        \toprule
        \rowcolor{color3} Guidance & PSNR $\uparrow$ & LPIPS $\downarrow$ & CLIP-IQA $\uparrow$ & DOVER $\uparrow$ \\
        \midrule
        Text & 24.87 & 0.2609 & 0.4002 & 0.7564 \\
        Video & \textbf{24.98} & 0.2744 & 0.3863 & 0.7335 \\
        Video\&Text & 24.92 & \textbf{0.2564} & \textbf{0.4086} & \textbf{0.7780} \\
        \bottomrule
\end{tabular}
\vspace{-1.mm}
\caption{Ablation on different guidance prompts.}
\vspace{-5.mm}
\label{tab:abl_guidance}
\end{wraptable}

\noindent \textbf{Video Representation Guidance.}
Different guidance prompts are compared in Tab.~\ref{tab:abl_guidance}. Text indicates the text embedding from a fixed text. Video denotes the video representation embedding from Eq.~\eqref{eq:video_prompt}. Video\&Text fuses both embeddings as in Eq.~\eqref{eq:video_text_prompt}. Video-only guidance noticeably degrades perceptual quality, likely due to distribution mismatch. Combining video and text embeddings balances fidelity and perceptual quality under the same backbone.

\begin{table*}[t]
    \centering
    \scriptsize
    \resizebox{\linewidth}{!}{
    \setlength{\tabcolsep}{1.5mm}
    \begin{tabular}{c | c  c | c c c c c c c c c c}
        \toprule
        \rowcolor{color3}  &  \multicolumn{2}{c|}{Methods} &  &  &  &  &  &  &  & \\
        \rowcolor{color3} \multirow{-2}{*}{Datasets} & VSR & VFI & \multirow{-2}{*}{PSNR $\uparrow$} & \multirow{-2}{*}{SSIM $\uparrow$} & \multirow{-2}{*}{LPIPS $\downarrow$} & \multirow{-2}{*}{DISTS $\downarrow$} & \multirow{-2}{*}{CLIP-IQA $\uparrow$} & \multirow{-2}{*}{MUSIQ $\uparrow$} & \multirow{-2}{*}{MANIQA $\uparrow$} & \multirow{-2}{*}{DOVER $\uparrow$} \\
        \midrule
        
        \multirow{12}{*}{UDM10} & 
        STAR~\cite{xie2025star} & BiM-VFI~\cite{seo2025bim} & \textcolor{blue}{23.09} & 0.6882 & 0.4054 & \textcolor{blue}{0.1888} & 0.2573 & 44.78 & 0.2330 & 0.5909 \\
        & STAR~\cite{xie2025star} & MoMo~\cite{lew2025disentangled} & 22.90 & \textcolor{blue}{0.6908} & 0.4030 & 0.1949 & 0.2690 & 42.89 & 0.2165 & 0.5358 \\
        & STAR~\cite{xie2025star} & TLBVFI~\cite{lyu2025tlb} & 22.73 & 0.6901 & \textcolor{blue}{0.4010} & 0.1947 & 0.2692 & 42.91 & 0.2154 & 0.5582 \\
        
        \cmidrule(lr){2-11}
        
        & SeedVR~\cite{wang2025seedvr} & BiM-VFI~\cite{seo2025bim} & 19.86 & 0.5470 & 0.4801 & 0.2099 & 0.3274 & \textcolor{blue}{50.96} & \textcolor{blue}{0.2587} & 0.5850 \\
        & SeedVR~\cite{wang2025seedvr} & MoMo~\cite{lew2025disentangled} & 19.79 & 0.5452 & 0.4783 & 0.2196 & 0.3471 & 49.52 & 0.2497 & 0.5723 \\
        & SeedVR~\cite{wang2025seedvr} & TLBVFI~\cite{lyu2025tlb} & 19.63 & 0.5481 & 0.4719 & 0.2187 & \textcolor{blue}{0.3513} & 49.48 & 0.2454 & 0.5825 \\
        
        \cmidrule(lr){2-11}
        
        & SeedVR2~\cite{wang2025seedvr2} & BiM-VFI~\cite{seo2025bim} & 21.05 & 0.6133 & 0.4427 & 0.2135 & 0.2431 & 40.12 & 0.1806 & 0.4536 \\
        & SeedVR2~\cite{wang2025seedvr2} & MoMo~\cite{lew2025disentangled} & 20.99 & 0.6124 & 0.4353 & 0.2170 & 0.2591 & 38.10 & 0.1644 & 0.4344 \\
        & SeedVR2~\cite{wang2025seedvr2} & TLBVFI~\cite{lyu2025tlb} & 20.80 & 0.6116 & 0.4319 & 0.2180 & 0.2628 & 38.61 & 0.1654 & 0.4567 \\
        
        \cmidrule(lr){2-11}
        
        & \multicolumn{2}{c|}{VEnhancer~\cite{he2024venhancer}} & 21.19 & 0.6692 & 0.4372 & 0.2198 & 0.2843 & 43.43 & 0.2128 & \textcolor{blue}{0.6063} \\
        \rowcolor{color-ours} \cellcolor{white} & \multicolumn{2}{c|}{DiffST (ours)}  & \textcolor{red}{24.92} & \textcolor{red}{0.7392} & \textcolor{red}{0.2564} & \textcolor{red}{0.1554} & \textcolor{red}{0.4086} & \textcolor{red}{62.47} & \textcolor{red}{0.3212} & \textcolor{red}{0.7780} \\
        
        \midrule[0.1em]
        
        \multirow{12}{*}{Vid4} & 
        STAR~\cite{xie2025star} & BiM-VFI~\cite{seo2025bim} & 17.82 & 0.4020 & 0.5551 & 0.2700 & \textcolor{red}{0.3570} & 46.34 & 0.2505 & 0.4299 \\
        & STAR~\cite{xie2025star} & MoMo~\cite{lew2025disentangled} & 17.73 & 0.4003 & 0.5568 & 0.2680 & \textcolor{blue}{0.3165} & 45.61 & 0.2577 & 0.3946 \\
        & STAR~\cite{xie2025star} & TLBVFI~\cite{lyu2025tlb} & 17.74 & 0.4033 & 0.5498 & 0.2654 & 0.3070 & 45.03 & 0.2524 & 0.3960 \\
        
        \cmidrule(lr){2-11}
        
        & SeedVR~\cite{wang2025seedvr} & BiM-VFI~\cite{seo2025bim} & 16.58 & 0.3372 & 0.4612 & 0.2150 & 0.3038 & \textcolor{blue}{63.73} & 0.3232 & \textcolor{blue}{0.5967} \\
        & SeedVR~\cite{wang2025seedvr} & MoMo~\cite{lew2025disentangled} & 16.47 & 0.3361 & 0.4646 & 0.2176 & 0.2941 & 63.11 & \textcolor{red}{0.3309} & 0.5621 \\
        & SeedVR~\cite{wang2025seedvr} & TLBVFI~\cite{lyu2025tlb} & 16.48 & 0.3387 & 0.4534 & 0.2145 & 0.2868 & 62.32 & 0.3172 & 0.5626 \\
        
        \cmidrule(lr){2-11}
        
        & SeedVR2~\cite{wang2025seedvr2} & BiM-VFI~\cite{seo2025bim} & \textcolor{blue}{18.77} & 0.4658 & 0.3594 & \textcolor{blue}{0.1996} & 0.2733 & 56.30 & 0.2472 & 0.4784 \\
        & SeedVR2~\cite{wang2025seedvr2} & MoMo~\cite{lew2025disentangled} & 18.64 & \textcolor{blue}{0.4664} & \textcolor{blue}{0.3593} & \textcolor{blue}{0.1996} & 0.2679 & 55.53 & 0.2508 & 0.4461 \\
        & SeedVR2~\cite{wang2025seedvr2} & TLBVFI~\cite{lyu2025tlb} & 18.59 & \textcolor{blue}{0.4664} & 0.3629 & 0.2011 & 0.2639 & 55.05 & 0.2496 & 0.4443 \\
        
        \cmidrule(lr){2-11}
        
        & \multicolumn{2}{c|}{VEnhancer~\cite{he2024venhancer}} & 15.98 & 0.3277 & 0.6404 & 0.3016 & 0.2507 & 37.46 & 0.2078 & 0.2914 \\
        \rowcolor{color-ours} \cellcolor{white} & \multicolumn{2}{c|}{DiffST (ours)}  & \textcolor{red}{19.99} & \textcolor{red}{0.5204} & \textcolor{red}{0.2699} & \textcolor{red}{0.1637} & 0.2735 & \textcolor{red}{66.12} & \textcolor{blue}{0.3254} & \textcolor{red}{0.6076} \\
        
        \midrule[0.1em]
        
        \multirow{12}{*}{MVSR4x} & 
        STAR~\cite{xie2025star} & BiM-VFI~\cite{seo2025bim} & 22.07 & 0.7343 & 0.4405 & 0.2636 & 0.2712 & 35.93 & 0.2931 & 0.2566 \\
        & STAR~\cite{xie2025star} & MoMo~\cite{lew2025disentangled} & 22.01 & 0.7344 & 0.4332 & 0.2636 & 0.2855 & 34.95 & 0.2908 & 0.2440 \\
        & STAR~\cite{xie2025star} & TLBVFI~\cite{lyu2025tlb} & 21.95 & 0.7349 & 0.4314 & 0.2612 & 0.2808 & 34.94 & 0.2907 & 0.2397 \\
        
        \cmidrule(lr){2-11}
        & SeedVR~\cite{wang2025seedvr} & BiM-VFI~\cite{seo2025bim} & 21.81 & 0.7252 & 0.4171 & 0.2299 & 0.2480 & \textcolor{blue}{38.77} & 0.2350 & \textcolor{blue}{0.3163} \\
        & SeedVR~\cite{wang2025seedvr} & MoMo~\cite{lew2025disentangled} & 21.80 & 0.7254 & 0.4123 & 0.2330 & 0.2908 & 38.07 & 0.2393 & 0.3063 \\
        & SeedVR~\cite{wang2025seedvr} & TLBVFI~\cite{lyu2025tlb} & 21.78 & 0.7262 & 0.4075 & 0.2319 & 0.2930 & 38.20 & 0.2391 & 0.3126 \\
        
        \cmidrule(lr){2-11}
        & SeedVR2~\cite{wang2025seedvr2} & BiM-VFI~\cite{seo2025bim} & 22.27 & \textcolor{red}{0.7657} & 0.3592 & \textcolor{blue}{0.2289} & 0.2117 & 32.65 & 0.2163 & 0.2401 \\
        & SeedVR2~\cite{wang2025seedvr2} & MoMo~\cite{lew2025disentangled} & \textcolor{red}{22.35} & \textcolor{blue}{0.7641} & 0.3566 & 0.2299 & 0.2377 & 31.76 & 0.2176 & 0.2369 \\
        & SeedVR2~\cite{wang2025seedvr2} & TLBVFI~\cite{lyu2025tlb} & \textcolor{blue}{22.30} & 0.7633 & \textcolor{blue}{0.3560} & 0.2295 & 0.2392 & 32.12 & 0.2191 & 0.2382 \\
        
        \cmidrule(lr){2-11}
        
        & \multicolumn{2}{c|}{VEnhancer~\cite{he2024venhancer}} & 20.37 & 0.7112 & 0.4562 & 0.2779 & \textcolor{blue}{0.2980} & 37.96 & \textcolor{blue}{0.3207} & 0.3064 \\
        \rowcolor{color-ours} \cellcolor{white} & \multicolumn{2}{c|}{DiffST (ours)}  & 22.24 & 0.7446 & \textcolor{red}{0.3320} & \textcolor{red}{0.2233} & \textcolor{red}{0.4565} & \textcolor{red}{60.99} & \textcolor{red}{0.3591} & \textcolor{red}{0.6739} \\
        
        \midrule[0.1em]
        
        \multirow{12}{*}{RealVSR} & 
        STAR~\cite{xie2025star} & BiM-VFI~\cite{seo2025bim} & 16.41 & 0.4607 & 0.3200 & 0.1654 & \textcolor{red}{0.5499} & \textcolor{blue}{73.21} & \textcolor{blue}{0.4296} & 0.7383 \\
        & STAR~\cite{xie2025star} & MoMo~\cite{lew2025disentangled} & 16.36 & 0.4604 & 0.3189 & 0.1596 & \textcolor{blue}{0.4919} & 72.65 & \textcolor{red}{0.4337} & 0.7155 \\
        & STAR~\cite{xie2025star} & TLBVFI~\cite{lyu2025tlb} & 16.48 & 0.4652 & 0.3192 & 0.1593 & 0.4722 & 71.85 & 0.4171 & 0.7158 \\
        
        \cmidrule(lr){2-11}
        
        & SeedVR~\cite{wang2025seedvr} & BiM-VFI~\cite{seo2025bim} & 17.71 & 0.4813 & 0.3232 & 0.1677 & 0.3314 & 62.10 & 0.3255 & 0.7023 \\
        & SeedVR~\cite{wang2025seedvr} & MoMo~\cite{lew2025disentangled} & 17.65 & 0.4794 & 0.3148 & 0.1641 & 0.3512 & 61.92 & 0.3300 & 0.6796 \\
        & SeedVR~\cite{wang2025seedvr} & TLBVFI~\cite{lyu2025tlb} & 17.58 & 0.4779 & 0.3165 & 0.1651 & 0.3514 & 61.25 & 0.3240 & 0.6845 \\
        
        \cmidrule(lr){2-11}
        
        & SeedVR2~\cite{wang2025seedvr2} & BiM-VFI~\cite{seo2025bim} & \textcolor{blue}{18.79} & 0.5552 & 0.2567 & 0.1365 & 0.3208 & 62.78 & 0.3324 & 0.6826 \\
        & SeedVR2~\cite{wang2025seedvr2} & MoMo~\cite{lew2025disentangled} & 18.70 & \textcolor{red}{0.5621} & \textcolor{blue}{0.2525} & \textcolor{blue}{0.1315} & 0.3373 & 61.76 & 0.3349 & 0.6632 \\
        & SeedVR2~\cite{wang2025seedvr2} & TLBVFI~\cite{lyu2025tlb} & 18.64 & \textcolor{blue}{0.5592} & 0.2588 & 0.1348 & 0.3318 & 60.98 & 0.3269 & 0.6638 \\
        
        \cmidrule(lr){2-11}
        
        & \multicolumn{2}{c|}{VEnhancer~\cite{he2024venhancer}} & 16.48 & 0.4274 & 0.3953 & 0.1755 & 0.3830 & 69.74 & 0.3827 & \textcolor{blue}{0.7395} \\
        \rowcolor{color-ours} \cellcolor{white} & \multicolumn{2}{c|}{DiffST (ours)}  & \textcolor{red}{19.01} & 0.5562 & \textcolor{red}{0.2151} & \textcolor{red}{0.1205} & 0.3833 & \textcolor{red}{74.95} & 0.4167 & \textcolor{red}{0.8048} \\
        
        \bottomrule
    \end{tabular}
    }
\vspace{-1.mm}
\caption{Quantitative results. The best and second best results are colored with \textcolor{red}{red} and \textcolor{blue}{blue}.}
\vspace{-4.mm}
\label{tab:quantitative}
\end{table*}

\begin{table}[t]
\scriptsize
\centering
\setlength{\tabcolsep}{1.mm}
\begin{tabular}{ll|ccc|ccc|ccc||cc} 
        \toprule
        \rowcolor{color3} & \textbf{VSR} & \multicolumn{3}{c|}{STAR} & \multicolumn{3}{c|}{SeedVR} & \multicolumn{3}{c||}{SeedVR2} & \multicolumn{2}{c}{\textbf{STVSR}} \\
        \rowcolor{color3} \cellcolor{color3}\multirow{-2}{*}{\textbf{Method}} & \textbf{VFI} & BiM-VFI & MoMo & TLBVFI & BiM-VFI & MoMo & TLBVFI & BiM-VFI & MoMo & TLBVFI & VEnhancer & DiffST \\
        \midrule
        \multicolumn{2}{l|}{Inference Step} & 15+1 & 15+8 & 15+10 & 50+1 & 50+8 & 50+10 & 1+1 & 1+8 & 1+10 & 15 & \textbf{1}\\
        \multicolumn{2}{l|}{Parameter (M)} & 2,499.78 & 2,566.48 & 2,539.60 & 3,404.26 & 3,470.96 & 3,444.08 & 3,404.26 & 3,470.96 & 3,444.08 & 2,496.59 & \textbf{1,581.46}\\
        \multicolumn{2}{l|}{Runtime (s)} & 48.42 & 51.69 & 58.32 & 90.21 & 93.49 & 100.11 & 22.99 & 26.27 & 32.89 & 124.29 & \textbf{7.12}\\
        \bottomrule
\end{tabular}
\vspace{1.mm}
\caption{Complexity comparison. We compare inference steps, parameter, and runtime. Runtime is measured using a 33-frame input video at a resolution of 720$\times$1280.}
\vspace{-10.mm}
\label{tab:time}
\end{table}

\vspace{-2.5mm}
\subsection{Comparison with State-of-the-Art Methods}
\label{sec:sota}
\vspace{-2.5mm}
We compare the proposed DiffST against several VSR+VFI approaches. The VSR methods contain: STAR~\cite{xie2025star}, SeedVR~\cite{wang2025seedvr}, and SeedVR2~\cite{wang2025seedvr2}, where SeedVR2 is a single-step diffusion model and the others are multi-step diffusion. For VFI, we include BiM-VFI~\cite{seo2025bim}, MoMo~\cite{lew2025disentangled}, and TLBVFI~\cite{lyu2025tlb}. BiM-VFI is an end-to-end model, while MoMo and TLBVFI are multi-step diffusion approaches. In addition, we compare with the multi-step diffusion-based STVSR method VEnhancer~\cite{he2024venhancer}.

\noindent \textbf{Quantitative Results.}
Table~\ref{tab:quantitative} summarizes the quantitative comparison, while Tab.~\ref{tab:time} reports the complexity comparison. Appendix~\ref{sec:app_more_metric_results} provides more temporal-consistency metrics, and Appendix~\ref{sec:app_more_stvsr} includes comparisons with additional single-stage STVSR methods. Our DiffST is competitive across synthetic and real-world datasets, ranking first or second on most metrics. For example, on the real-world dataset MVSR4x, compared to the method (SeedVR~\cite{wang2025seedvr}+BiM-VFI~\cite{seo2025bim}), it improves DOVER by approximately \textbf{94\%} on this benchmark.

Moreover, our method achieves lower parameter counts and lower computational complexity. Meanwhile, thanks to its single-step, video-level, one-stage STVSR design, DiffST runs significantly faster. Compared with the two-stage pipeline (SeedVR~\cite{wang2025seedvr}+TLBVFI~\cite{lyu2025tlb}), our method achieves a speedup of \textbf{14$\times$}. It also outperforms the single-stage diffusion-based VEnhancer~\cite{he2024venhancer} with a speedup of \textbf{17$\times$}.

\begin{figure*}[!t]
\scriptsize
\centering
\begin{tabular}{cccccccc}

\hspace{-0.48cm}
\begin{adjustbox}{valign=t}
\begin{tabular}{c}
\includegraphics[width=0.2215\textwidth]{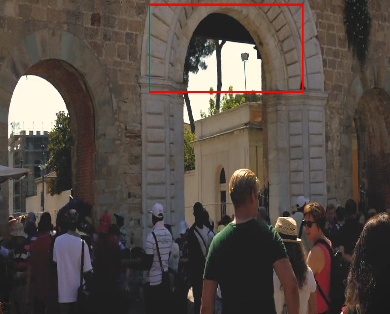}
\\
UDM10: 000
\end{tabular}
\end{adjustbox}
\hspace{-0.46cm}
\begin{adjustbox}{valign=t}
\begin{tabular}{cccccc}
\includegraphics[width=0.190\textwidth]{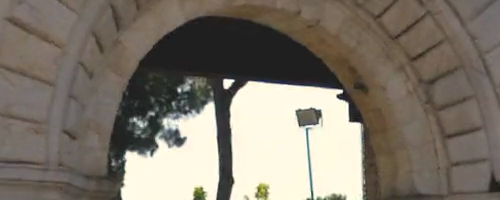} \hspace{-4.mm} &
\includegraphics[width=0.190\textwidth]{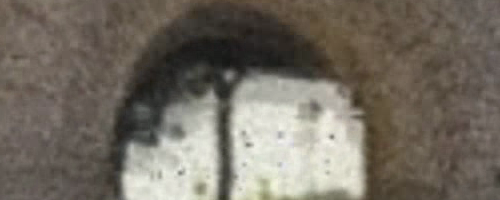} \hspace{-4.mm} &
\includegraphics[width=0.190\textwidth]{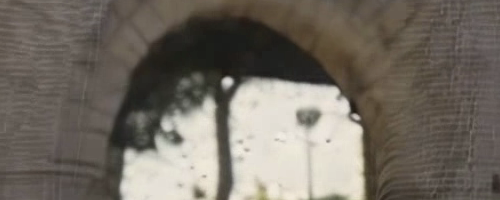} \hspace{-4.mm} &
\includegraphics[width=0.190\textwidth]{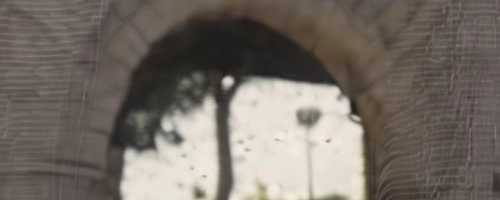} \hspace{-4.mm} &
\\ 
HR \hspace{-4.mm} &
LR \hspace{-4.mm} &
STAR+TLBVFI \hspace{-4.mm} &
STAR+BiM-VFI \hspace{-4.mm} &
\\
\includegraphics[width=0.190\textwidth]{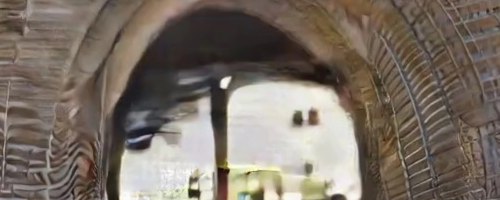} \hspace{-4.mm} &
\includegraphics[width=0.190\textwidth]{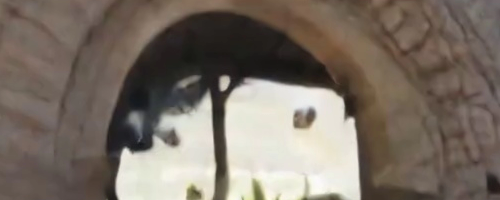} \hspace{-4.mm} &
\includegraphics[width=0.190\textwidth]{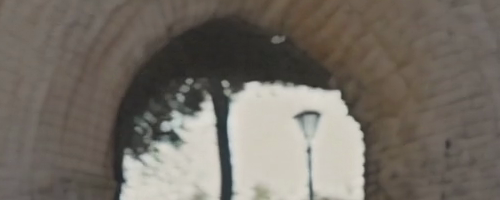} \hspace{-4.mm} &
\includegraphics[width=0.190\textwidth]{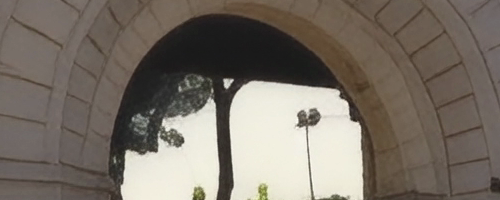} \hspace{-4.mm} &
\\ 
SeedVR+BiM-VFI \hspace{-4.mm} &
SeedVR2+BiM-VFI \hspace{-4.mm} &
VEnhancer \hspace{-4.mm} &
DiffST (ours) \hspace{-4mm}
\\
\end{tabular}
\end{adjustbox}
\\

\hspace{-0.48cm}
\begin{adjustbox}{valign=t}
\begin{tabular}{c}
\includegraphics[width=0.2215\textwidth]{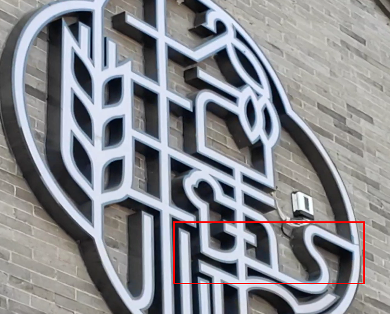}
\\
MVSR4x: 465
\end{tabular}
\end{adjustbox}
\hspace{-0.46cm}
\begin{adjustbox}{valign=t}
\begin{tabular}{cccccc}
\includegraphics[width=0.190\textwidth]{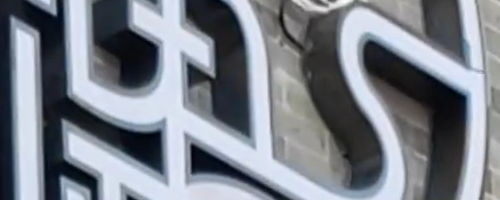} \hspace{-4.mm} &
\includegraphics[width=0.190\textwidth]{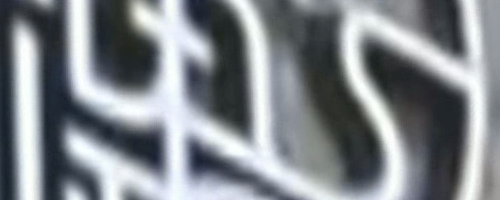} \hspace{-4.mm} &
\includegraphics[width=0.190\textwidth]{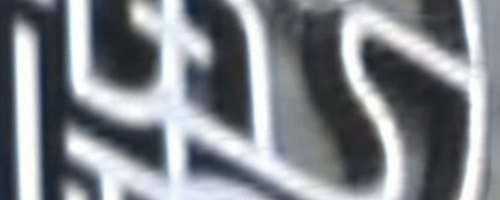} \hspace{-4.mm} &
\includegraphics[width=0.190\textwidth]{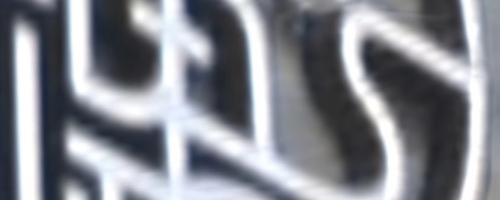} \hspace{-4.mm} &
\\ 
HR \hspace{-4.mm} &
LR \hspace{-4.mm} &
STAR+TLBVFI \hspace{-4.mm} &
STAR+BiM-VFI \hspace{-4.mm} &
\\
\includegraphics[width=0.190\textwidth]{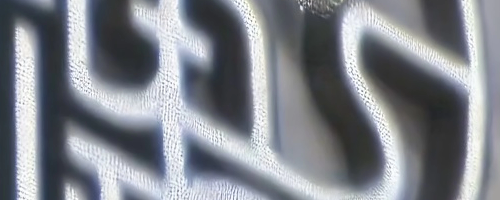} \hspace{-4.mm} &
\includegraphics[width=0.190\textwidth]{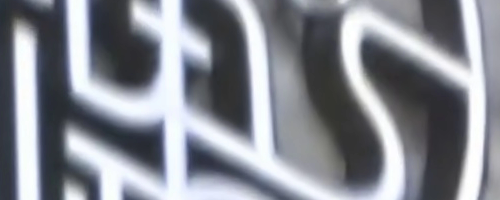} \hspace{-4.mm} &
\includegraphics[width=0.190\textwidth]{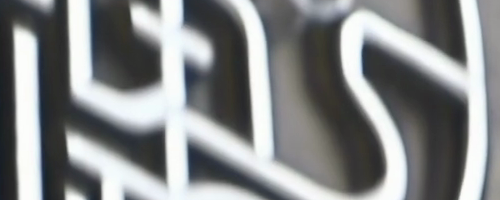} \hspace{-4.mm} &
\includegraphics[width=0.190\textwidth]{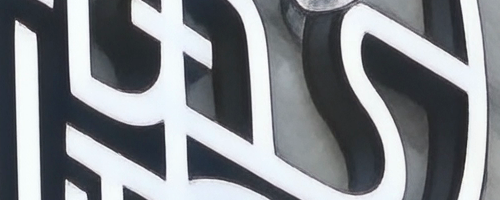} \hspace{-4.mm} &
\\ 
SeedVR+BiM-VFI \hspace{-4.mm} &
SeedVR2+BiM-VFI \hspace{-4.mm} &
VEnhancer \hspace{-4.mm} &
DiffST (ours) \hspace{-4mm}
\\
\end{tabular}
\end{adjustbox}
\\

\hspace{-0.48cm}
\begin{adjustbox}{valign=t}
\begin{tabular}{c}
\includegraphics[width=0.2215\textwidth]{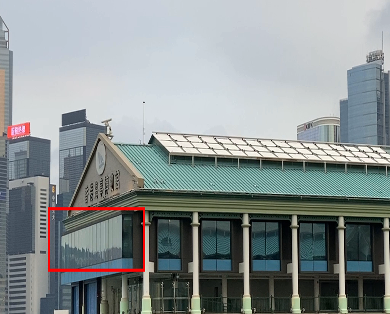}
\\
RealVSR: 044
\end{tabular}
\end{adjustbox}
\hspace{-0.46cm}
\begin{adjustbox}{valign=t}
\begin{tabular}{cccccc}
\includegraphics[width=0.190\textwidth]{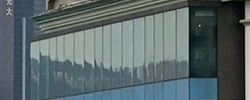} \hspace{-4.mm} &
\includegraphics[width=0.190\textwidth]{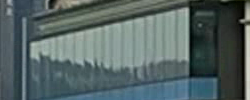} \hspace{-4.mm} &
\includegraphics[width=0.190\textwidth]{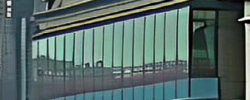} \hspace{-4.mm} &
\includegraphics[width=0.190\textwidth]{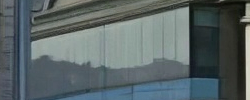} \hspace{-4.mm} &
\\ 
HR \hspace{-4.mm} &
LR \hspace{-4.mm} &
STAR+TLBVFI \hspace{-4.mm} &
STAR+BiM-VFI \hspace{-4.mm} &
\\
\includegraphics[width=0.190\textwidth]{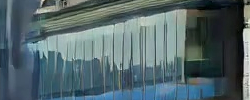} \hspace{-4.mm} &
\includegraphics[width=0.190\textwidth]{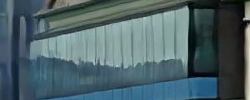} \hspace{-4.mm} &
\includegraphics[width=0.190\textwidth]{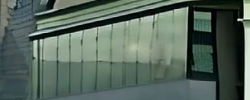} \hspace{-4.mm} &
\includegraphics[width=0.190\textwidth]{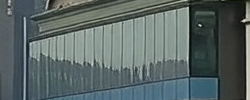} \hspace{-4.mm} &
\\ 
SeedVR+BiM-VFI \hspace{-4.mm} &
SeedVR2+BiM-VFI \hspace{-4.mm} &
VEnhancer \hspace{-4.mm} &
DiffST (ours) \hspace{-4mm}
\\
\end{tabular}
\end{adjustbox}

\end{tabular}
\vspace{-1.mm}
\caption{Qualitative results on synthetic (UDM10~\cite{tao2017detail}) and real-world (MVSR4x~\cite{wang2023benchmark} and RealVSR~\cite{yang2021real}) benchmarks. Our method achieves impressive performance.}
\vspace{-3.mm}
\label{fig:visual}
\end{figure*}

\begin{figure*}[t]
    \centering
    \includegraphics[width=\linewidth]{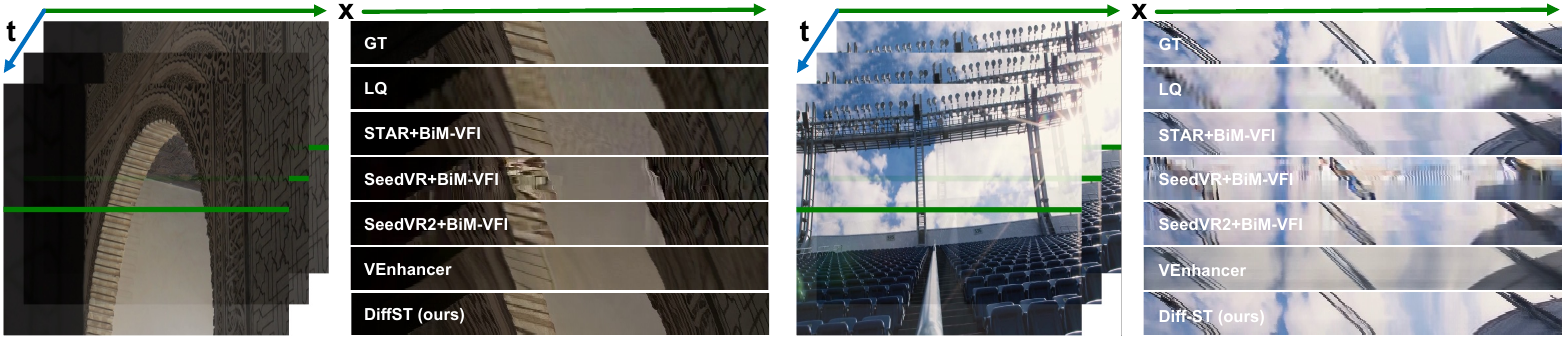}
    \vspace{-4.mm}
    \caption{Consistency comparison with other STVSR methods. We stack the green dots on each frame along the temporal axis. Our method produces richer details and smoother transitions.}
    \vspace{-4.mm}
    \label{fig:consistency}
\end{figure*}

\noindent \textbf{Qualitative Results.}
Figure~\ref{fig:visual} shows qualitative comparisons for challenging cases. Against existing SOTA methods, our approach recovers more realistic and sharper details. For example, in the first case, competing methods introduce severe artifacts or excessive blur, while our DiffST successfully restores the fine textures on the stone gate. These examples verify the effectiveness of our approach. They also agree with the quantitative results in Tab.~\ref{tab:quantitative}. More visual examples appear in Appendix~\ref{sec:app_more_qualitative}.

\noindent \textbf{Temporal Consistency.}
Temporal consistency is visualized in Fig.~\ref{fig:consistency}. The compared methods exhibit noticeable temporal instability. Specifically, two-stage pipelines introduce significant artifacts. For instance, in the left case, in the third case, strong noise appears in the blank region between objects. The single-stage method VEnhancer also generates unrealistic and inconsistent content. In contrast, our DiffST produces smoother and more coherent frame-to-frame transitions. 

\vspace{-2.mm}
\section{Conclusion}
\vspace{-2.mm}
In this paper, we propose DiffST, an efficient video diffusion model for real-world space–time video super-resolution (STVSR). DiffST performs video-level one-step sampling to ensure efficient inference. Meanwhile, we introduce cross-frame context aggregation (CFCA) and video representation guidance (VRG) to enhance the utilization of spatiotemporal information and improve reconstruction quality. CFCA aggregates multiple keyframes to better construct intermediate frames. VRG extracts the video representation with rich spatiotemporal information to guide the diffusion generation process. Experiments on synthetic and real-world benchmarks reveal that our proposed DiffST achieves impressive space-time video super-resolution performance and high efficiency.

{\small
\bibliographystyle{plain}
\bibliography{example_paper}

\begin{thebibliography}{10}

\bibitem{cao2022towards}
Jiezhang Cao, Jingyun Liang, Kai Zhang, Wenguan Wang, Qin Wang, Yulun Zhang,
  Hao Tang, and Luc Van~Gool.
\newblock Towards interpretable video super-resolution via alternating
  optimization.
\newblock In {\em ECCV}, 2022.

\bibitem{chan2021basicvsr}
Kelvin~CK Chan, Xintao Wang, Ke~Yu, Chao Dong, and Chen~Change Loy.
\newblock Basicvsr: The search for essential components in video
  super-resolution and beyond.
\newblock In {\em CVPR}, 2021.

\bibitem{chan2022basicvsr++}
Kelvin~CK Chan, Shangchen Zhou, Xiangyu Xu, and Chen~Change Loy.
\newblock Basicvsr++: Improving video super-resolution with enhanced
  propagation and alignment.
\newblock In {\em CVPR}, 2022.

\bibitem{chan2022investigating}
Kelvin~CK Chan, Shangchen Zhou, Xiangyu Xu, and Chen~Change Loy.
\newblock Investigating tradeoffs in real-world video super-resolution.
\newblock In {\em CVPR}, 2022.

\bibitem{chen2023motif}
Yi-Hsin Chen, Si-Cun Chen, Yen-Yu Lin, and Wen-Hsiao Peng.
\newblock Motif: Learning motion trajectories with local implicit neural
  functions for continuous space-time video super-resolution.
\newblock In {\em ICCV}, 2023.

\bibitem{chen2022videoinr}
Zeyuan Chen, Yinbo Chen, Jingwen Liu, Xingqian Xu, Vidit Goel, Zhangyang Wang,
  Humphrey Shi, and Xiaolong Wang.
\newblock Videoinr: Learning video implicit neural representation for
  continuous space-time super-resolution.
\newblock In {\em CVPR}, 2022.

\bibitem{chen2025dove}
Zheng Chen, Zichen Zou, Kewei Zhang, Xiongfei Su, Xin Yuan, Yong Guo, and Yulun
  Zhang.
\newblock Dove: Efficient one-step diffusion model for real-world video
  super-resolution.
\newblock In {\em NeurIPS}, 2025.

\bibitem{danier2022flolpips}
Duolikun Danier, Fan Zhang, and David Bull.
\newblock Flolpips: A bespoke video quality metric for frame interpolation.
\newblock In {\em PCS}, 2022.

\bibitem{danier2024ldmvfi}
Duolikun Danier, Fan Zhang, and David Bull.
\newblock Ldmvfi: Video frame interpolation with latent diffusion models.
\newblock In {\em AAAI}, 2024.

\bibitem{ding2020image}
Keyan Ding, Kede Ma, Shiqi Wang, and Eero~P Simoncelli.
\newblock Image quality assessment: Unifying structure and texture similarity.
\newblock {\em TPAMI}, 2020.

\bibitem{ding2021cdfi}
Tianyu Ding, Luming Liang, Zhihui Zhu, and Ilya Zharkov.
\newblock Cdfi: Compression-driven network design for frame interpolation.
\newblock In {\em CVPR}, 2021.

\bibitem{geng2022rstt}
Zhicheng Geng, Luming Liang, Tianyu Ding, and Ilya Zharkov.
\newblock Rstt: Real-time spatial temporal transformer for space-time video
  super-resolution.
\newblock In {\em CVPR}, 2022.

\bibitem{haris2020space}
Muhammad Haris, Greg Shakhnarovich, and Norimichi Ukita.
\newblock Space-time-aware multi-resolution video enhancement.
\newblock In {\em CVPR}, 2020.

\bibitem{he2024venhancer}
Jingwen He, Tianfan Xue, Dongyang Liu, Xinqi Lin, Peng Gao, Dahua Lin, Yu~Qiao,
  Wanli Ouyang, and Ziwei Liu.
\newblock Venhancer: Generative space-time enhancement for video generation.
\newblock {\em arXiv preprint arXiv:2407.07667}, 2024.

\bibitem{hu2023cycmunet+}
Mengshun Hu, Kui Jiang, Zheng Wang, Xiang Bai, and Ruimin Hu.
\newblock Cycmunet+: Cycle-projected mutual learning for spatial-temporal video
  super-resolution.
\newblock {\em TPAMI}, 2023.

\bibitem{huang2022real}
Zhewei Huang, Tianyuan Zhang, Wen Heng, Boxin Shi, and Shuchang Zhou.
\newblock Real-time intermediate flow estimation for video frame interpolation.
\newblock In {\em ECCV}, 2022.

\bibitem{jin2023unified}
Xin Jin, Longhai Wu, Jie Chen, Youxin Chen, Jayoon Koo, and Cheul-hee Hahm.
\newblock A unified pyramid recurrent network for video frame interpolation.
\newblock In {\em CVPR}, 2023.

\bibitem{jo2018deep}
Younghyun Jo, Seoung~Wug Oh, Jaeyeon Kang, and Seon~Joo Kim.
\newblock Deep video super-resolution network using dynamic upsampling filters
  without explicit motion compensation.
\newblock In {\em CVPR}, 2018.

\bibitem{kappeler2016video}
Armin Kappeler, Seunghwan Yoo, Qiqin Dai, and Aggelos~K Katsaggelos.
\newblock Video super-resolution with convolutional neural networks.
\newblock {\em TCI}, 2016.

\bibitem{ke2021musiq}
Junjie Ke, Qifei Wang, Yilin Wang, Peyman Milanfar, and Feng Yang.
\newblock Musiq: Multi-scale image quality transformer.
\newblock In {\em ICCV}, 2021.

\bibitem{kim2025bf}
Eunjin Kim, Hyeonjin Kim, Kyong~Hwan Jin, and Jaejun Yoo.
\newblock Bf-stvsr: B-splines and fourier---best friends for high fidelity
  spatial-temporal video super-resolution.
\newblock In {\em CVPR}, 2025.

\bibitem{kong2024hunyuanvideo}
Weijie Kong, Qi~Tian, Zijian Zhang, Rox Min, Zuozhuo Dai, Jin Zhou, Jiangfeng
  Xiong, Xin Li, Bo~Wu, Jianwei Zhang, et~al.
\newblock Hunyuanvideo: A systematic framework for large video generative
  models.
\newblock {\em arXiv preprint arXiv:2412.03603}, 2024.

\bibitem{lee2020adacof}
Hyeongmin Lee, Taeoh Kim, Tae-young Chung, Daehyun Pak, Yuseok Ban, and
  Sangyoun Lee.
\newblock Adacof: Adaptive collaboration of flows for video frame
  interpolation.
\newblock In {\em CVPR}, 2020.

\bibitem{lew2025disentangled}
Jaihyun Lew, Jooyoung Choi, Chaehun Shin, Dahuin Jung, and Sungroh Yoon.
\newblock Disentangled motion modeling for video frame interpolation.
\newblock In {\em AAAI}, 2025.

\bibitem{li2025asymmetric}
Jianze Li, Yong Guo, Yulun Zhang, and Xiaokang Yang.
\newblock Asymmetric vae for one-step video super-resolution acceleration.
\newblock {\em arXiv preprint arXiv:2509.24142}, 2025.

\bibitem{li2025diffvsr}
Xiaohui Li, Yihao Liu, Shuo Cao, Ziyan Chen, Shaobin Zhuang, Xiangyu Chen,
  Yinan He, Yi~Wang, and Yu~Qiao.
\newblock Diffvsr: Enhancing real-world video super-resolution with diffusion
  models for advanced visual quality and temporal consistency.
\newblock {\em arXiv preprint arXiv:2501.10110}, 2025.

\bibitem{liao2015video}
Renjie Liao, Xin Tao, Ruiyu Li, Ziyang Ma, and Jiaya Jia.
\newblock Video super-resolution via deep draft-ensemble learning.
\newblock In {\em ICCV}, 2015.

\bibitem{liu2011bayesian}
Ce~Liu and Deqing Sun.
\newblock A bayesian approach to adaptive video super resolution.
\newblock In {\em CVPR}, 2011.

\bibitem{liu2017robust}
Ding Liu, Zhaowen Wang, Yuchen Fan, Xianming Liu, Zhangyang Wang, Shiyu Chang,
  and Thomas Huang.
\newblock Robust video super-resolution with learned temporal dynamics.
\newblock In {\em ICCV}, 2017.

\bibitem{liu2017video}
Ziwei Liu, Raymond~A Yeh, Xiaoou Tang, Yiming Liu, and Aseem Agarwala.
\newblock Video frame synthesis using deep voxel flow.
\newblock In {\em ICCV}, 2017.

\bibitem{lyu2025tlb}
Zonglin Lyu and Chen Chen.
\newblock Tlb-vfi: Temporal-aware latent brownian bridge diffusion for video
  frame interpolation.
\newblock In {\em ICCV}, 2025.

\bibitem{mudenagudi2010space}
Uma Mudenagudi, Subhashis Banerjee, and Prem~Kumar Kalra.
\newblock Space-time super-resolution using graph-cut optimization.
\newblock {\em TPAMI}, 2010.

\bibitem{niklaus2018context}
Simon Niklaus and Feng Liu.
\newblock Context-aware synthesis for video frame interpolation.
\newblock In {\em CVPR}, 2018.

\bibitem{seo2025bim}
Wonyong Seo, Jihyong Oh, and Munchurl Kim.
\newblock Bim-vfi: Bidirectional motion field-guided frame interpolation for
  video with non-uniform motions.
\newblock In {\em CVPR}, 2025.

\bibitem{shechtman2002increasing}
Eli Shechtman, Yaron Caspi, and Michal Irani.
\newblock Increasing space-time resolution in video.
\newblock In {\em ECCV}, 2002.

\bibitem{shi2022rethinking}
Shuwei Shi, Jinjin Gu, Liangbin Xie, Xintao Wang, Yujiu Yang, and Chao Dong.
\newblock Rethinking alignment in video super-resolution transformers.
\newblock In {\em NeurIPS}, 2022.

\bibitem{tao2017detail}
Xin Tao, Hongyun Gao, Renjie Liao, Jue Wang, and Jiaya Jia.
\newblock Detail-revealing deep video super-resolution.
\newblock In {\em ICCV}, 2017.

\bibitem{tian2020tdan}
Yapeng Tian, Yulun Zhang, Yun Fu, and Chenliang Xu.
\newblock Tdan: Temporally-deformable alignment network for video
  super-resolution.
\newblock In {\em CVPR}, 2020.

\bibitem{wan2025wan}
Team Wan, Ang Wang, Baole Ai, Bin Wen, Chaojie Mao, Chen-Wei Xie, Di~Chen,
  Feiwu Yu, Haiming Zhao, Jianxiao Yang, et~al.
\newblock Wan: Open and advanced large-scale video generative models.
\newblock {\em arXiv preprint arXiv:2503.20314}, 2025.

\bibitem{wang2023exploring}
Jianyi Wang, Kelvin~CK Chan, and Chen~Change Loy.
\newblock Exploring clip for assessing the look and feel of images.
\newblock In {\em AAAI}, 2023.

\bibitem{wang2025seedvr2}
Jianyi Wang, Shanchuan Lin, Zhijie Lin, Yuxi Ren, Meng Wei, Zongsheng Yue,
  Shangchen Zhou, Hao Chen, Yang Zhao, Ceyuan Yang, et~al.
\newblock Seedvr2: One-step video restoration via diffusion adversarial
  post-training.
\newblock {\em arXiv preprint arXiv:2506.05301}, 2025.

\bibitem{wang2025seedvr}
Jianyi Wang, Zhijie Lin, Meng Wei, Yang Zhao, Ceyuan Yang, Fei Xiao,
  Chen~Change Loy, and Lu~Jiang.
\newblock Seedvr: Seeding infinity in diffusion transformer towards generic
  video restoration.
\newblock In {\em CVPR}, 2025.

\bibitem{wang2023benchmark}
Ruohao Wang, Xiaohui Liu, Zhilu Zhang, Xiaohe Wu, Chun-Mei Feng, Lei Zhang, and
  Wangmeng Zuo.
\newblock Benchmark dataset and effective inter-frame alignment for real-world
  video super-resolution.
\newblock In {\em CVPRW}, 2023.

\bibitem{wang2024generative}
Xiaojuan Wang, Boyang Zhou, Brian Curless, Ira Kemelmacher-Shlizerman,
  Aleksander Holynski, and Steven~M Seitz.
\newblock Generative inbetweening: Adapting image-to-video models for keyframe
  interpolation.
\newblock In {\em ICLR}, 2025.

\bibitem{wang2004image}
Zhou Wang, Alan~C Bovik, Hamid~R Sheikh, and Eero~P Simoncelli.
\newblock Image quality assessment: from error visibility to structural
  similarity.
\newblock {\em TIP}, 2004.

\bibitem{wei2025evenhancer}
Shuoyan Wei, Feng Li, Shengeng Tang, Yao Zhao, and Huihui Bai.
\newblock Evenhancer: Empowering effectiveness, efficiency and generalizability
  for continuous space-time video super-resolution with events.
\newblock In {\em CVPR}, 2025.

\bibitem{wu2023exploring}
Haoning Wu, Erli Zhang, Liang Liao, Chaofeng Chen, Jingwen Hou, Annan Wang,
  Wenxiu Sun, Qiong Yan, and Weisi Lin.
\newblock Exploring video quality assessment on user generated contents from
  aesthetic and technical perspectives.
\newblock In {\em ICCV}, 2023.

\bibitem{wu2023seesr}
Rongyuan Wu, Tao Yang, Lingchen Sun, Zhengqiang Zhang, Shuai Li, and Lei Zhang.
\newblock Seesr: Towards semantics-aware real-world image super-resolution.
\newblock In {\em CVPR}, 2024.

\bibitem{xiang2020zooming}
Xiaoyu Xiang, Yapeng Tian, Yulun Zhang, Yun Fu, Jan~P Allebach, and Chenliang
  Xu.
\newblock Zooming slow-mo: Fast and accurate one-stage space-time video
  super-resolution.
\newblock In {\em CVPR}, 2020.

\bibitem{xie2025star}
Rui Xie, Yinhong Liu, Penghao Zhou, Chen Zhao, Jun Zhou, Kai Zhang, Zhenyu
  Zhang, Jian Yang, Zhenheng Yang, and Ying Tai.
\newblock Star: Spatial-temporal augmentation with text-to-video models for
  real-world video super-resolution.
\newblock {\em arXiv preprint arXiv:2501.02976}, 2025.

\bibitem{xu2021temporal}
Gang Xu, Jun Xu, Zhen Li, Liang Wang, Xing Sun, and Ming-Ming Cheng.
\newblock Temporal modulation network for controllable space-time video
  super-resolution.
\newblock In {\em CVPR}, 2021.

\bibitem{xu2019deep}
Rui Xu, Xiaoxiao Li, Bolei Zhou, and Chen~Change Loy.
\newblock Deep flow-guided video inpainting.
\newblock In {\em CVPR}, 2019.

\bibitem{yang2024vibidsampler}
Serin Yang, Taesung Kwon, and Jong~Chul Ye.
\newblock Vibidsampler: Enhancing video interpolation using bidirectional
  diffusion sampler.
\newblock In {\em ICLR}, 2025.

\bibitem{yang2022maniqa}
Sidi Yang, Tianhe Wu, Shuwei Shi, Shanshan Lao, Yuan Gong, Mingdeng Cao, Jiahao
  Wang, and Yujiu Yang.
\newblock Maniqa: Multi-dimension attention network for no-reference image
  quality assessment.
\newblock In {\em CVPRW}, 2022.

\bibitem{yang2024motion}
Xi~Yang, Chenhang He, Jianqi Ma, and Lei Zhang.
\newblock Motion-guided latent diffusion for temporally consistent real-world
  video super-resolution.
\newblock In {\em ECCV}, 2024.

\bibitem{yang2021real}
Xi~Yang, Wangmeng Xiang, Hui Zeng, and Lei Zhang.
\newblock Real-world video super-resolution: A benchmark dataset and a
  decomposition based learning scheme.
\newblock In {\em CVPR}, 2021.

\bibitem{yang2024cogvideox}
Zhuoyi Yang, Jiayan Teng, Wendi Zheng, Ming Ding, Shiyu Huang, Jiazheng Xu,
  Yuanming Yang, Wenyi Hong, Xiaohan Zhang, Guanyu Feng, et~al.
\newblock Cogvideox: Text-to-video diffusion models with an expert transformer.
\newblock In {\em ICLR}, 2025.

\bibitem{zhang2023adding}
Lvmin Zhang, Anyi Rao, and Maneesh Agrawala.
\newblock Adding conditional control to text-to-image diffusion models.
\newblock In {\em ICCV}, 2023.

\bibitem{zhang2018unreasonable}
Richard Zhang, Phillip Isola, Alexei~A Efros, Eli Shechtman, and Oliver Wang.
\newblock The unreasonable effectiveness of deep features as a perceptual
  metric.
\newblock In {\em CVPR}, 2018.

\bibitem{zhang2025eden}
Zihao Zhang, Haoran Chen, Haoyu Zhao, Guansong Lu, Yanwei Fu, Hang Xu, and
  Zuxuan Wu.
\newblock Eden: Enhanced diffusion for high-quality large-motion video frame
  interpolation.
\newblock In {\em CVPR}, 2025.

\bibitem{zhang2025infvsr}
Ziqing Zhang, Kai Liu, Zheng Chen, Xi~Li, Yucong Chen, Bingnan Duan, Linghe
  Kong, and Yulun Zhang.
\newblock Infvsr: Breaking length limits of generic video super-resolution.
\newblock {\em arXiv preprint arXiv:2510.00948}, 2025.

\bibitem{zhou2023exploring}
Kun Zhou, Wenbo Li, Xiaoguang Han, and Jiangbo Lu.
\newblock Exploring motion ambiguity and alignment for high-quality video frame
  interpolation.
\newblock In {\em CVPR}, 2023.

\bibitem{zhou2023propainter}
Shangchen Zhou, Chongyi Li, Kelvin~CK Chan, and Chen~Change Loy.
\newblock Propainter: Improving propagation and transformer for video
  inpainting.
\newblock In {\em ICCV}, 2023.

\bibitem{zhou2024upscale}
Shangchen Zhou, Peiqing Yang, Jianyi Wang, Yihang Luo, and Chen~Change Loy.
\newblock Upscale-a-video: Temporal-consistent diffusion model for real-world
  video super-resolution.
\newblock In {\em CVPR}, 2024.

\bibitem{zhuang2025flashvsr}
Junhao Zhuang, Shi Guo, Xin Cai, Xiaohui Li, Yihao Liu, Chun Yuan, and Tianfan
  Xue.
\newblock Flashvsr: Towards real-time diffusion-based streaming video
  super-resolution.
\newblock {\em arXiv preprint arXiv:2510.12747}, 2025.

\end{thebibliography}
}


\appendix

\section{Evaluation on More Metrics}
\label{sec:app_more_metrics}
In this section, we introduce more metrics, particularly those designed for assessing temporal consistency, to provide a more comprehensive analysis of STVSR performance. This appendix complements the evaluation metrics and main quantitative comparison in Sec.~\ref{sec:setting} and Sec.~\ref{sec:sota}.

\subsection{Metrics}
\label{sec:app_more_metric_def}
We incorporate \textbf{tOF/tLP} to measure temporal consistency, and adopt \textbf{FloLPIPS}~\cite{danier2022flolpips} to evaluate perceptual video quality. 
The detailed is described as follows.

\noindent \textbf{tOF/tLP.}
The tOF and tLP evaluate the temporal consistency between generated and ground-truth videos by comparing frame-to-frame changes over time.

tOF measures the difference in pixel-level motion between the output and ground-truth using optical flow estimated from consecutive frames to reflect motion consistency over time.

tLP measures temporal perceptual discrepancies using deep feature maps. The metrics are defined as:
\begin{equation}
\begin{aligned}
\mathrm{tOF}&=\Vert OF(\mathbf{I}_{gt}^{(t-1)},\mathbf{I}_{gt}^{(t)})-OF(\mathbf{I}_{out}^{(t-1)},\mathbf{I}_{out}^{(t)})\Vert_1,\\
\mathrm{tLP}&=\Vert LP(\mathbf{I}_{gt}^{(t-1)},\mathbf{I}_{gt}^{(t)})-LP(\mathbf{I}_{out}^{(t-1)},\mathbf{I}_{out}^{(t)})\Vert_1,
\end{aligned}
\end{equation}
where, $OF$ denotes optical-flow estimation, $LP$ denotes LPIPS-based perceptual features, $\mathbf{I}_{gt}$ denotes round-truth frames, and $\mathbf{I}_{out}$ is the model output video at adjacent timesteps.
For both metrics, lower values indicate better temporal consistency and fewer temporal artifacts.

\noindent \textbf{FloLPIPS.}
This metric~\cite{danier2022flolpips} is a full-reference perceptual video quality metric. It is based on LPIPS~\cite{zhang2018unreasonable} with motion-aware modeling to better assess perceptual degradation in videos over temporal changes.
It incorporates motion information derived from optical flow to capture distortions that arise across consecutive frames. By integrating both appearance differences and motion-related deviations, FloLPIPS provides a perceptually aligned assessment of overall video quality for restored sequences.
For this metric, lower values indicate better perceptual quality.

\subsection{Results}
\label{sec:app_more_metric_results}

\begin{wraptable}{r}{0.60\textwidth}
\scriptsize
\centering
\vspace{-11.mm}
\setlength{\tabcolsep}{1.mm}
\begin{tabular}{cc|ccc|ccc} 
\toprule[0.15em]
\rowcolor{color3} \multicolumn{2}{c|}{Methods}  & \multicolumn{3}{c|}{UDM10} & \multicolumn{3}{c}{RealVSR}\\
\rowcolor{color3} VSR & VFI & {tOF $\downarrow$} & {tLP $\downarrow$} & {FloLPIPS $\downarrow$} & {tOF $\downarrow$} & {tLP $\downarrow$} & {FloLPIPS $\downarrow$} \\

\midrule[0.15em]
STAR & BiM-VFI & 1.59 & 2.08 & 0.4017 & 1.75 & \textcolor{blue}{2.91} & 0.2878 \\
STAR & MoMo & \textcolor{blue}{1.51} & 2.27 & 0.4075 & \textcolor{blue}{1.63} & 3.11 & 0.3203 \\
STAR & TLBVFI & 1.71 & \textcolor{red}{1.86} & \textcolor{black}{0.4094} & 1.79 & 2.93 & \textcolor{black}{0.3233} \\

\midrule

SeedVR & BiM-VFI & 2.16 & 4.45 & 0.4882 & 1.96 & 4.20 & 0.3220 \\
SeedVR & MoMo & 1.99 & 2.49 & 0.4570 & 1.83 & 3.78 & 0.3180 \\
SeedVR & TLBVFI & 2.93 & 3.63 & 0.4608 & 2.00 & 3.57 & 0.3215 \\

\midrule
SeedVR2 & BiM-VFI & 1.53 & 2.44 & \textcolor{blue}{0.3753} & 1.69 & 3.41 & 0.2529 \\
SeedVR2 & MoMo & 1.65 & 2.50 & 0.4948 & 1.75 & 3.35 & \textcolor{blue}{0.2541} \\
SeedVR2 & TLBVFI & 2.35 & 3.45 & 0.4891 & 1.64 & 3.54 & \textcolor{black}{0.2633} \\

\midrule

\multicolumn{2}{c|}{VEnhancer} & 1.63 & 2.29 & 0.4119 & 1.79 & 3.42 & 0.3908 \\
\rowcolor{color-ours} \multicolumn{2}{c|}{DiffST (ours)} & \textcolor{red}{1.25} & \textcolor{blue}{1.89} & \textcolor{red}{0.2649} & \textcolor{red}{1.56} & \textcolor{red}{2.89} & \textcolor{red}{0.2147} \\
\bottomrule[0.15em]
\end{tabular}
\vspace{-0.5mm}
\caption{Quantitative comparison on more metrics.}
\vspace{-10.mm}
\label{tab:more_metric}
\end{wraptable}

We evaluate tOF/tLP and FloLPIPS on the synthetic UDM10 dataset and the real-world RealVSR dataset. The results are presented in Tab~\ref{tab:more_metric}.
Compared with existing STVSR methods and VSR+VFI pipelines, our proposed DiffST achieves significantly better temporal consistency, reflected by lower tOF and tLP scores. In addition, DiffST outperforms competing approaches on the perceptual video metric FloLPIPS.

\section{Comparison with More STVSR Methods}
\label{sec:app_more_stvsr}
We compare our DiffST with more single-stage space-time video super-resolution methods, including Zooming Slow-Mo~\cite{xiang2020zooming} and BF-STVSR~\cite{kim2025bf}. This appendix extends the main comparison with state-of-the-art methods in Sec.~\ref{sec:sota} using additional one-stage baselines.

\noindent \textbf{Zooming Slow-Mo.}
Zooming Slow-Mo is a single-stage STVSR method that captures local temporal information through a feature-level temporal interpolation network. It employs a deformable ConvLSTM to jointly align and aggregate cross-frame features, followed by a reconstruction network that generates high-quality outputs from the aggregated representations.

\noindent \textbf{BF-STVSR.}
BF-STVSR is a continuous space–time video super-resolution framework that improves reconstruction quality by explicitly modeling spatial and temporal features. It uses a B-spline mapper to achieve smooth temporal interpolation, enhancing spatial detail and temporal consistency.

\noindent \textbf{Results.}
The comparison results are shown in Tab.~\ref{tab:more_method}. Our method outperforms the STVSR baselines on both synthetic and real-world datasets, with particularly strong gains in perceptual metrics such as CLIP-IQA and DOVER. These findings demonstrate the effectiveness of our approach.

\section{More Qualitative Results}
\label{sec:app_more_qualitative}
\begin{figure*}[!t]
\vspace{20.mm}
\scriptsize
\centering
\begin{tabular}{cccccccc}

\hspace{-0.48cm}
\begin{adjustbox}{valign=t}
\begin{tabular}{c}
\includegraphics[width=0.2215\textwidth]{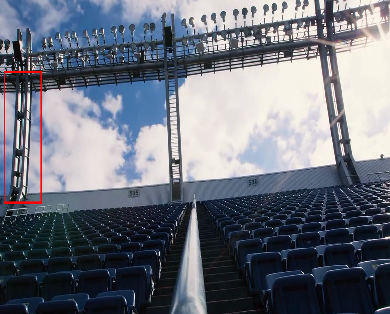}
\\
UDM10: 002
\end{tabular}
\end{adjustbox}
\hspace{-0.46cm}
\begin{adjustbox}{valign=t}
\begin{tabular}{cccccc}
\includegraphics[width=0.190\textwidth]{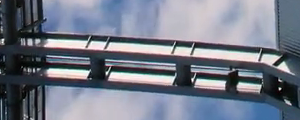} \hspace{-4.mm} &
\includegraphics[width=0.190\textwidth]{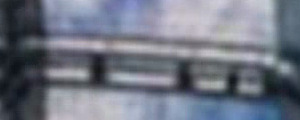} \hspace{-4.mm} &
\includegraphics[width=0.190\textwidth]{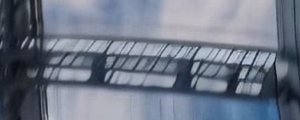} \hspace{-4.mm} &
\includegraphics[width=0.190\textwidth]{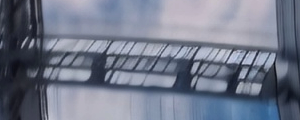} \hspace{-4.mm} &
\\ 
HR \hspace{-4.mm} &
LR \hspace{-4.mm} &
STAR+TLBVFI \hspace{-4.mm} &
STAR+BiM-VFI \hspace{-4.mm} &
\\
\includegraphics[width=0.190\textwidth]{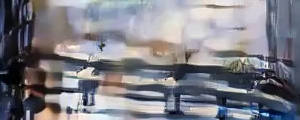} \hspace{-4.mm} &
\includegraphics[width=0.190\textwidth]{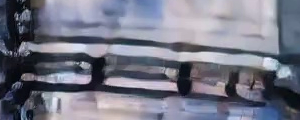} \hspace{-4.mm} &
\includegraphics[width=0.190\textwidth]{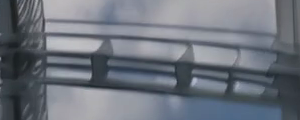} \hspace{-4.mm} &
\includegraphics[width=0.190\textwidth]{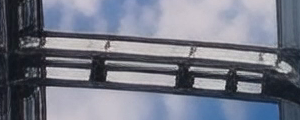} \hspace{-4.mm} &
\\ 
SeedVR+BiM-VFI \hspace{-4.mm} &
SeedVR2+BiM-VFI \hspace{-4.mm} &
VEnhancer \hspace{-4.mm} &
DiffST (ours) \hspace{-4mm}
\\
\end{tabular}
\end{adjustbox}
\\

\hspace{-0.48cm}
\begin{adjustbox}{valign=t}
\begin{tabular}{c}
\includegraphics[width=0.2215\textwidth]{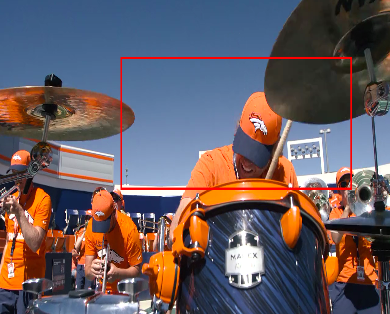}
\\
UDM10: 003
\end{tabular}
\end{adjustbox}
\hspace{-0.46cm}
\begin{adjustbox}{valign=t}
\begin{tabular}{cccccc}
\includegraphics[width=0.190\textwidth]{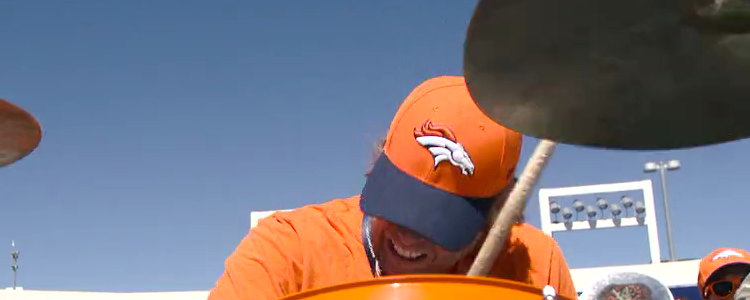} \hspace{-4.mm} &
\includegraphics[width=0.190\textwidth]{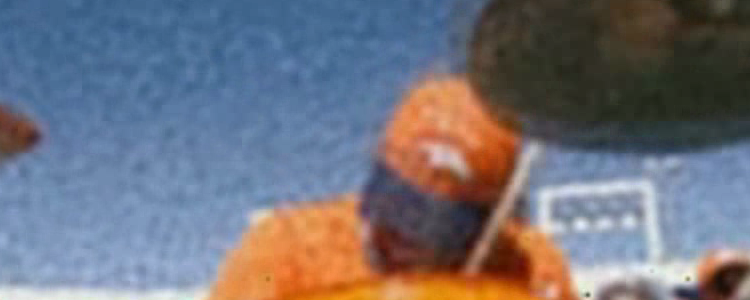} \hspace{-4.mm} &
\includegraphics[width=0.190\textwidth]{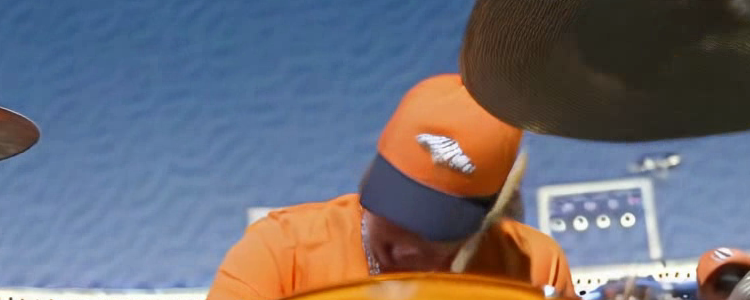} \hspace{-4.mm} &
\includegraphics[width=0.190\textwidth]{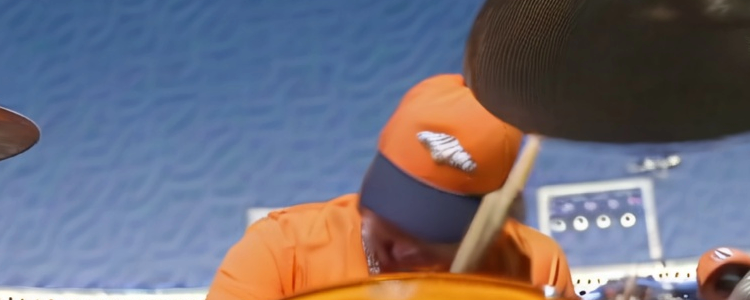} \hspace{-4.mm} &
\\ 
HR \hspace{-4.mm} &
LR \hspace{-4.mm} &
STAR+TLBVFI \hspace{-4.mm} &
STAR+BiM-VFI \hspace{-4.mm} &
\\
\includegraphics[width=0.190\textwidth]{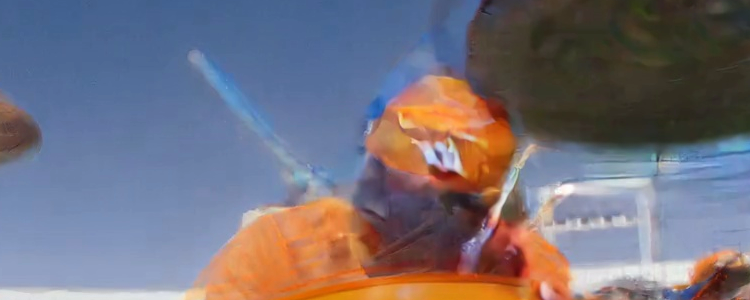} \hspace{-4.mm} &
\includegraphics[width=0.190\textwidth]{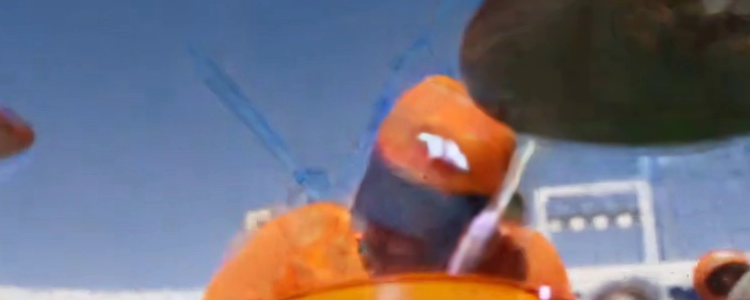} \hspace{-4.mm} &
\includegraphics[width=0.190\textwidth]{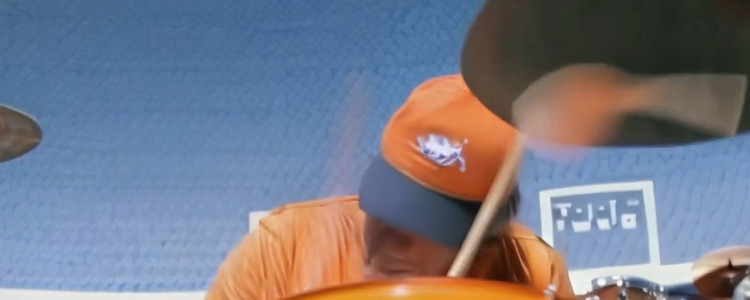} \hspace{-4.mm} &
\includegraphics[width=0.190\textwidth]{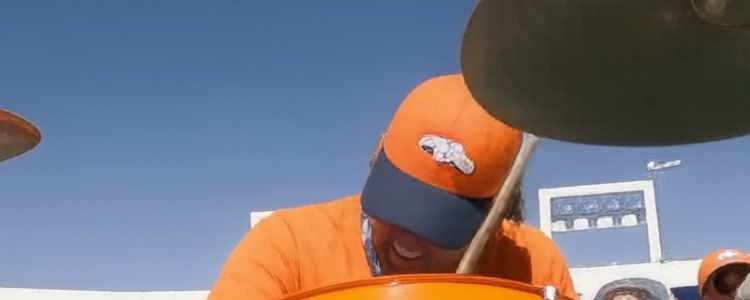} \hspace{-4.mm} &
\\ 
SeedVR+BiM-VFI \hspace{-4.mm} &
SeedVR2+BiM-VFI \hspace{-4.mm} &
VEnhancer \hspace{-4.mm} &
DiffST (ours) \hspace{-4mm}
\\
\end{tabular}
\end{adjustbox}
\\

\hspace{-0.48cm}
\begin{adjustbox}{valign=t}
\begin{tabular}{c}
\includegraphics[width=0.2215\textwidth]{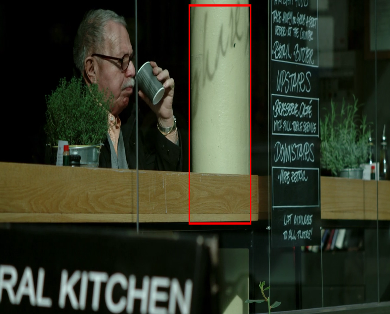}
\\
UDM10: 004
\end{tabular}
\end{adjustbox}
\hspace{-0.46cm}
\begin{adjustbox}{valign=t}
\begin{tabular}{cccccc}
\includegraphics[width=0.190\textwidth]{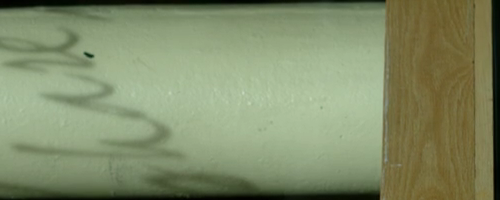} \hspace{-4.mm} &
\includegraphics[width=0.190\textwidth]{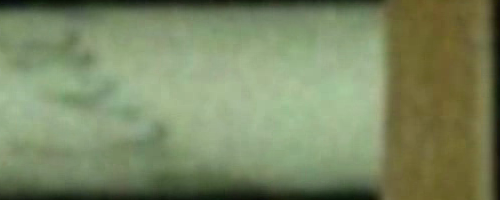} \hspace{-4.mm} &
\includegraphics[width=0.190\textwidth]{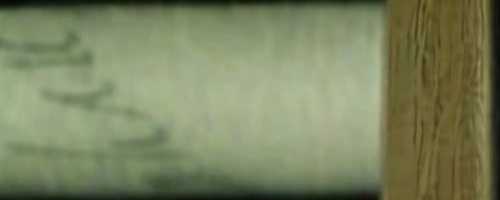} \hspace{-4.mm} &
\includegraphics[width=0.190\textwidth]{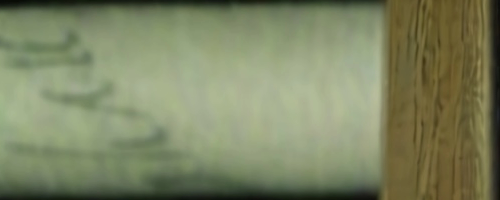} \hspace{-4.mm} &
\\ 
HR \hspace{-4.mm} &
LR \hspace{-4.mm} &
STAR+TLBVFI \hspace{-4.mm} &
STAR+BiM-VFI \hspace{-4.mm} &
\\
\includegraphics[width=0.190\textwidth]{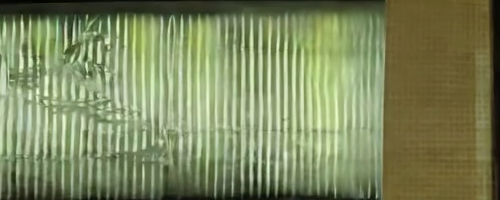} \hspace{-4.mm} &
\includegraphics[width=0.190\textwidth]{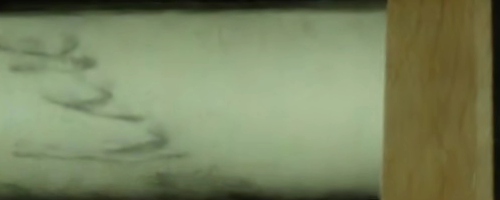} \hspace{-4.mm} &
\includegraphics[width=0.190\textwidth]{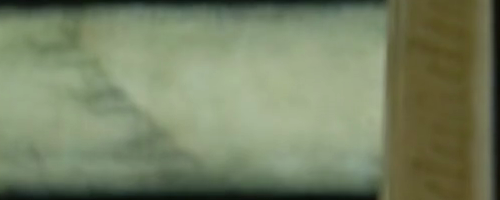} \hspace{-4.mm} &
\includegraphics[width=0.190\textwidth]{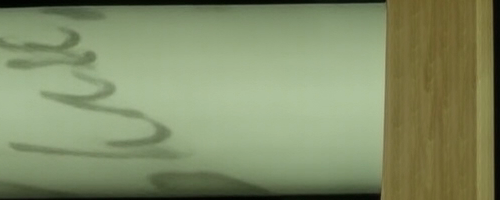} \hspace{-4.mm} &
\\ 
SeedVR+BiM-VFI \hspace{-4.mm} &
SeedVR2+BiM-VFI \hspace{-4.mm} &
VEnhancer \hspace{-4.mm} &
DiffST (ours) \hspace{-4mm}
\\
\end{tabular}
\end{adjustbox}
\\

\hspace{-0.48cm}
\begin{adjustbox}{valign=t}
\begin{tabular}{c}
\includegraphics[width=0.2215\textwidth]{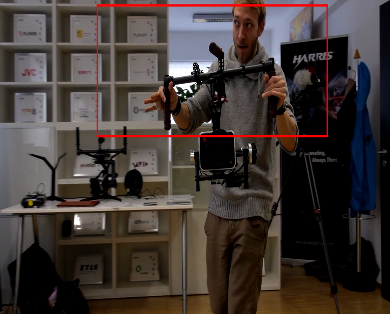}
\\
UDM10: 008
\end{tabular}
\end{adjustbox}
\hspace{-0.46cm}
\begin{adjustbox}{valign=t}
\begin{tabular}{cccccc}
\includegraphics[width=0.190\textwidth]{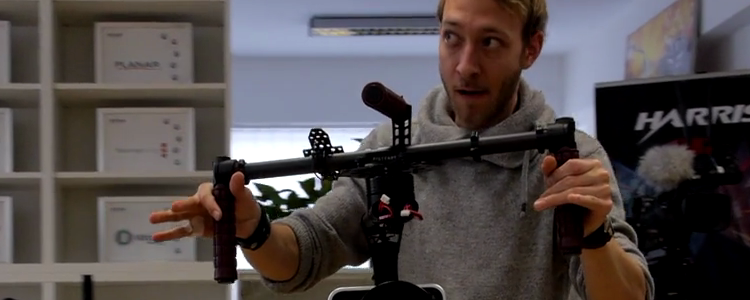} \hspace{-4.mm} &
\includegraphics[width=0.190\textwidth]{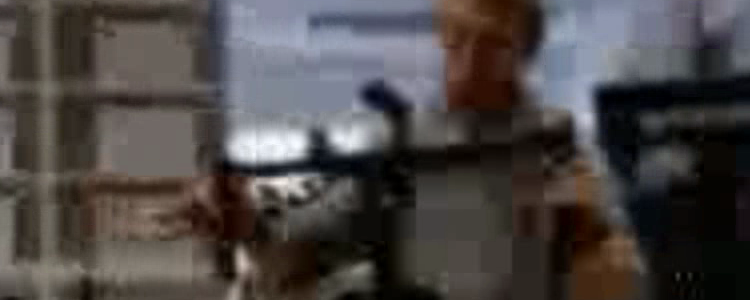} \hspace{-4.mm} &
\includegraphics[width=0.190\textwidth]{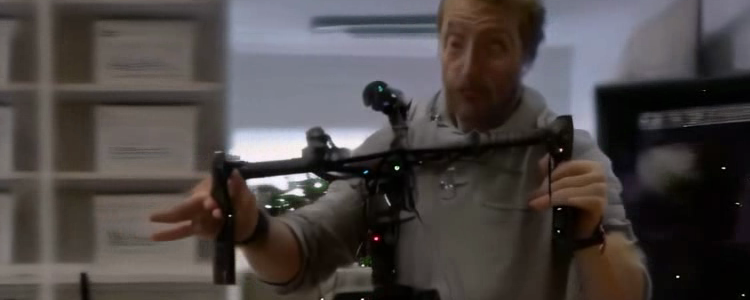} \hspace{-4.mm} &
\includegraphics[width=0.190\textwidth]{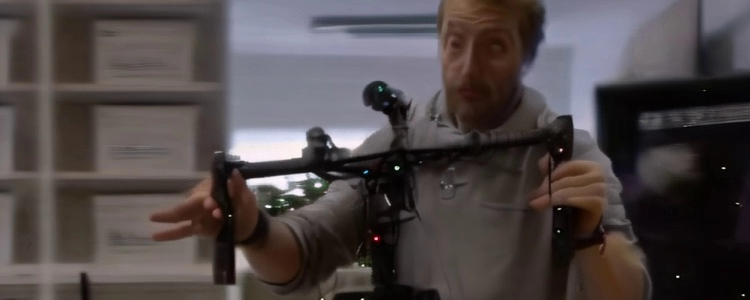} \hspace{-4.mm} &
\\ 
HR \hspace{-4.mm} &
LR \hspace{-4.mm} &
STAR+TLBVFI \hspace{-4.mm} &
STAR+BiM-VFI \hspace{-4.mm} &
\\
\includegraphics[width=0.190\textwidth]{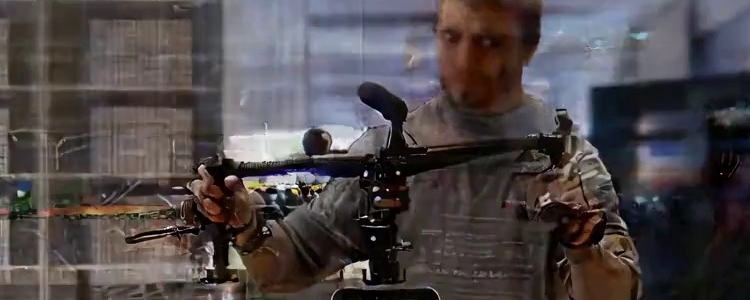} \hspace{-4.mm} &
\includegraphics[width=0.190\textwidth]{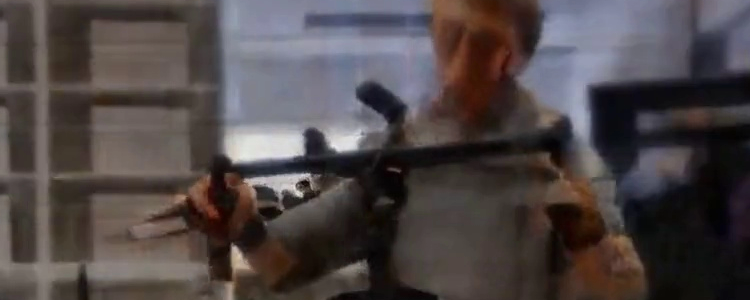} \hspace{-4.mm} &
\includegraphics[width=0.190\textwidth]{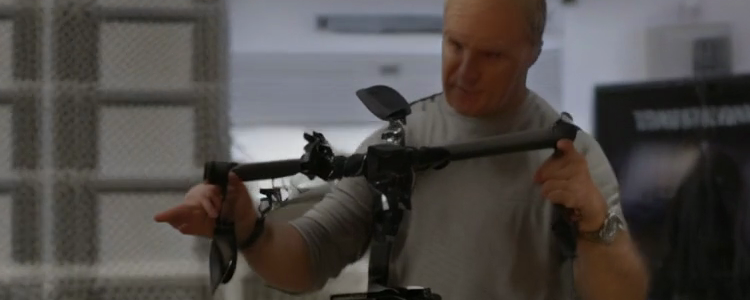} \hspace{-4.mm} &
\includegraphics[width=0.190\textwidth]{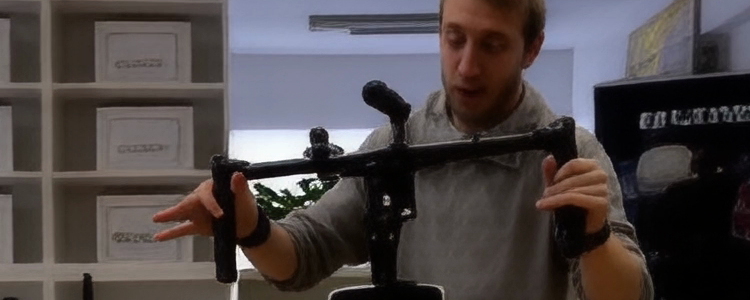} \hspace{-4.mm} &
\\ 
SeedVR+BiM-VFI \hspace{-4.mm} &
SeedVR2+BiM-VFI \hspace{-4.mm} &
VEnhancer \hspace{-4.mm} &
DiffST (ours) \hspace{-4mm}
\\
\end{tabular}
\end{adjustbox}
\\

\hspace{-0.48cm}
\begin{adjustbox}{valign=t}
\begin{tabular}{c}
\includegraphics[width=0.2215\textwidth]{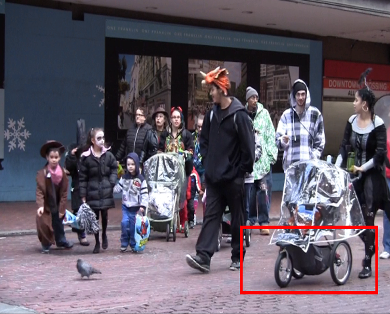}
\\
Vid4: walk
\end{tabular}
\end{adjustbox}
\hspace{-0.46cm}
\begin{adjustbox}{valign=t}
\begin{tabular}{cccccc}
\includegraphics[width=0.190\textwidth]{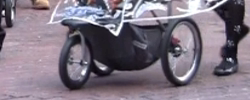} \hspace{-4.mm} &
\includegraphics[width=0.190\textwidth]{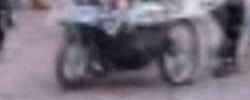} \hspace{-4.mm} &
\includegraphics[width=0.190\textwidth]{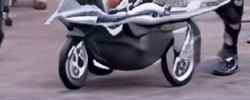} \hspace{-4.mm} &
\includegraphics[width=0.190\textwidth]{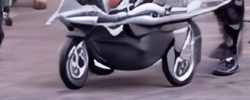} \hspace{-4.mm} &
\\ 
HR \hspace{-4.mm} &
LR \hspace{-4.mm} &
STAR+TLBVFI \hspace{-4.mm} &
STAR+BiM-VFI \hspace{-4.mm} &
\\
\includegraphics[width=0.190\textwidth]{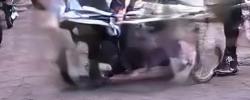} \hspace{-4.mm} &
\includegraphics[width=0.190\textwidth]{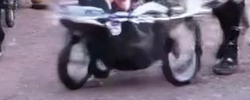} \hspace{-4.mm} &
\includegraphics[width=0.190\textwidth]{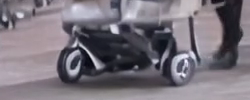} \hspace{-4.mm} &
\includegraphics[width=0.190\textwidth]{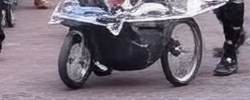} \hspace{-4.mm} &
\\ 
SeedVR+BiM-VFI \hspace{-4.mm} &
SeedVR2+BiM-VFI \hspace{-4.mm} &
VEnhancer \hspace{-4.mm} &
DiffST (ours) \hspace{-4mm}
\\
\end{tabular}
\end{adjustbox}
\\

\hspace{-0.48cm}
\begin{adjustbox}{valign=t}
\begin{tabular}{c}
\includegraphics[width=0.2215\textwidth]{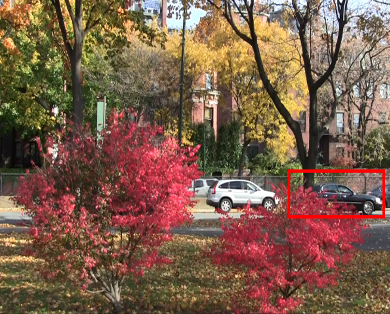}
\\
Vid4: foliage
\end{tabular}
\end{adjustbox}
\hspace{-0.46cm}
\begin{adjustbox}{valign=t}
\begin{tabular}{cccccc}
\includegraphics[width=0.190\textwidth]{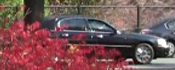} \hspace{-4.mm} &
\includegraphics[width=0.190\textwidth]{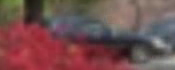} \hspace{-4.mm} &
\includegraphics[width=0.190\textwidth]{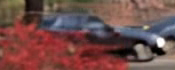} \hspace{-4.mm} &
\includegraphics[width=0.190\textwidth]{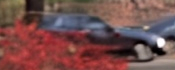} \hspace{-4.mm} &
\\ 
HR \hspace{-4.mm} &
LR \hspace{-4.mm} &
STAR+TLBVFI \hspace{-4.mm} &
STAR+BiM-VFI \hspace{-4.mm} &
\\
\includegraphics[width=0.190\textwidth]{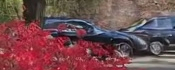} \hspace{-4.mm} &
\includegraphics[width=0.190\textwidth]{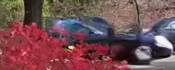} \hspace{-4.mm} &
\includegraphics[width=0.190\textwidth]{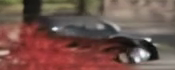} \hspace{-4.mm} &
\includegraphics[width=0.190\textwidth]{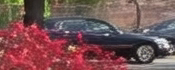} \hspace{-4.mm} &
\\ 
SeedVR+BiM-VFI \hspace{-4.mm} &
SeedVR2+BiM-VFI \hspace{-4.mm} &
VEnhancer \hspace{-4.mm} &
DiffST (ours) \hspace{-4mm}
\\
\end{tabular}
\end{adjustbox}

\end{tabular}
\caption{Visual comparison on synthetic (UDM10~\cite{tao2017detail} and Vid4~\cite{liu2011bayesian}) datasets.}
\vspace{20.mm}
\label{fig:visual-supp-1}
\end{figure*}

\begin{figure*}[!t]
\vspace{20.mm}
\scriptsize
\centering
\begin{tabular}{cccccccc}

\hspace{-0.48cm}
\begin{adjustbox}{valign=t}
\begin{tabular}{c}
\includegraphics[width=0.2215\textwidth]{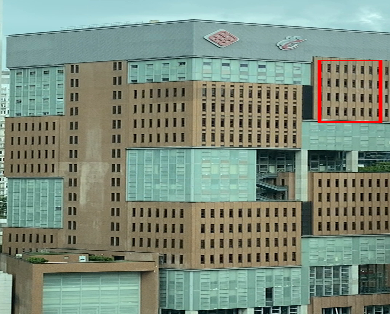}
\\
RealVSR: 018
\end{tabular}
\end{adjustbox}
\hspace{-0.46cm}
\begin{adjustbox}{valign=t}
\begin{tabular}{cccccc}
\includegraphics[width=0.190\textwidth]{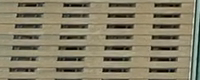} \hspace{-4.mm} &
\includegraphics[width=0.190\textwidth]{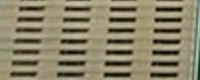} \hspace{-4.mm} &
\includegraphics[width=0.190\textwidth]{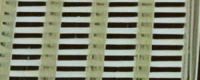} \hspace{-4.mm} &
\includegraphics[width=0.190\textwidth]{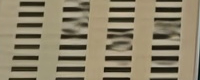} \hspace{-4.mm} &
\\ 
HR \hspace{-4.mm} &
LR \hspace{-4.mm} &
STAR+TLBVFI \hspace{-4.mm} &
STAR+BiM-VFI \hspace{-4.mm} &
\\
\includegraphics[width=0.190\textwidth]{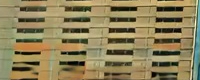} \hspace{-4.mm} &
\includegraphics[width=0.190\textwidth]{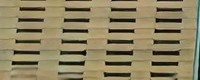} \hspace{-4.mm} &
\includegraphics[width=0.190\textwidth]{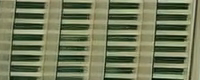} \hspace{-4.mm} &
\includegraphics[width=0.190\textwidth]{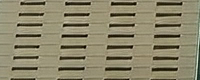} \hspace{-4.mm} &
\\ 
SeedVR+BiM-VFI \hspace{-4.mm} &
SeedVR2+BiM-VFI \hspace{-4.mm} &
VEnhancer \hspace{-4.mm} &
DiffST (ours) \hspace{-4mm}
\\
\end{tabular}
\end{adjustbox}
\\

\hspace{-0.48cm}
\begin{adjustbox}{valign=t}
\begin{tabular}{c}
\includegraphics[width=0.2215\textwidth]{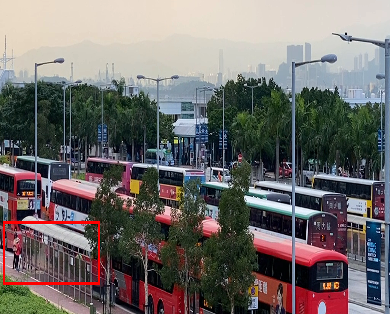}
\\
RealVSR: 039
\end{tabular}
\end{adjustbox}
\hspace{-0.46cm}
\begin{adjustbox}{valign=t}
\begin{tabular}{cccccc}
\includegraphics[width=0.190\textwidth]{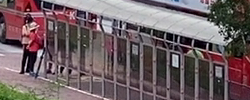} \hspace{-4.mm} &
\includegraphics[width=0.190\textwidth]{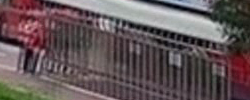} \hspace{-4.mm} &
\includegraphics[width=0.190\textwidth]{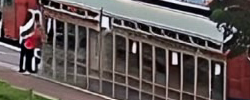} \hspace{-4.mm} &
\includegraphics[width=0.190\textwidth]{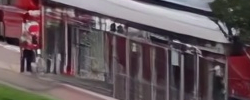} \hspace{-4.mm} &
\\ 
HR \hspace{-4.mm} &
LR \hspace{-4.mm} &
STAR+TLBVFI \hspace{-4.mm} &
STAR+BiM-VFI \hspace{-4.mm} &
\\
\includegraphics[width=0.190\textwidth]{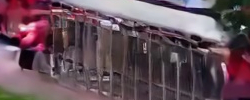} \hspace{-4.mm} &
\includegraphics[width=0.190\textwidth]{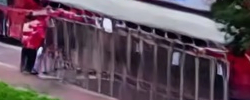} \hspace{-4.mm} &
\includegraphics[width=0.190\textwidth]{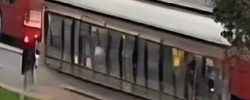} \hspace{-4.mm} &
\includegraphics[width=0.190\textwidth]{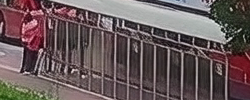} \hspace{-4.mm} &
\\ 
SeedVR+BiM-VFI \hspace{-4.mm} &
SeedVR2+BiM-VFI \hspace{-4.mm} &
VEnhancer \hspace{-4.mm} &
DiffST (ours) \hspace{-4mm}
\\
\end{tabular}
\end{adjustbox}
\\

\hspace{-0.48cm}
\begin{adjustbox}{valign=t}
\begin{tabular}{c}
\includegraphics[width=0.2215\textwidth]{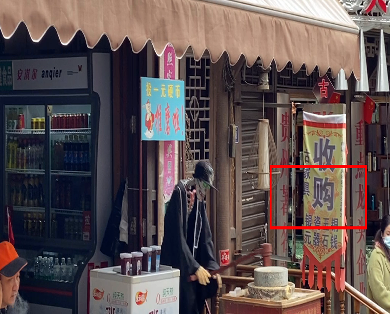}
\\
RealVSR: 170
\end{tabular}
\end{adjustbox}
\hspace{-0.46cm}
\begin{adjustbox}{valign=t}
\begin{tabular}{cccccc}
\includegraphics[width=0.190\textwidth]{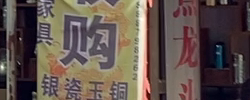} \hspace{-4.mm} &
\includegraphics[width=0.190\textwidth]{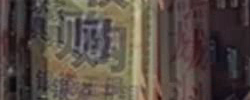} \hspace{-4.mm} &
\includegraphics[width=0.190\textwidth]{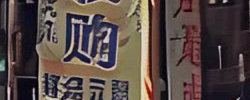} \hspace{-4.mm} &
\includegraphics[width=0.190\textwidth]{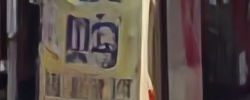} \hspace{-4.mm} &
\\ 
HR \hspace{-4.mm} &
LR \hspace{-4.mm} &
STAR+TLBVFI \hspace{-4.mm} &
STAR+BiM-VFI \hspace{-4.mm} &
\\
\includegraphics[width=0.190\textwidth]{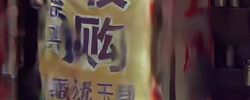} \hspace{-4.mm} &
\includegraphics[width=0.190\textwidth]{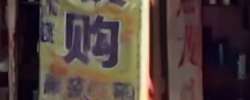} \hspace{-4.mm} &
\includegraphics[width=0.190\textwidth]{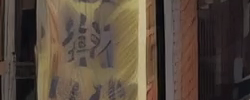} \hspace{-4.mm} &
\includegraphics[width=0.190\textwidth]{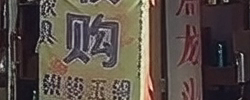} \hspace{-4.mm} &
\\ 
SeedVR+BiM-VFI \hspace{-4.mm} &
SeedVR2+BiM-VFI \hspace{-4.mm} &
VEnhancer \hspace{-4.mm} &
DiffST (ours) \hspace{-4mm}
\\
\end{tabular}
\end{adjustbox}
\\

\hspace{-0.48cm}
\begin{adjustbox}{valign=t}
\begin{tabular}{c}
\includegraphics[width=0.2215\textwidth]{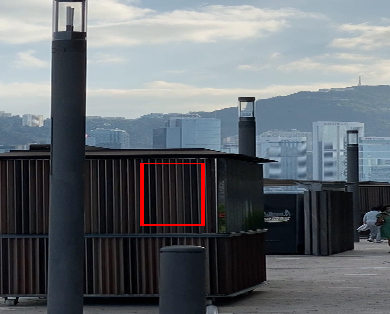}
\\
RealVSR: 211
\end{tabular}
\end{adjustbox}
\hspace{-0.46cm}
\begin{adjustbox}{valign=t}
\begin{tabular}{cccccc}
\includegraphics[width=0.190\textwidth]{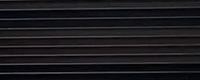} \hspace{-4.mm} &
\includegraphics[width=0.190\textwidth]{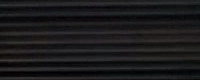} \hspace{-4.mm} &
\includegraphics[width=0.190\textwidth]{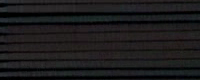} \hspace{-4.mm} &
\includegraphics[width=0.190\textwidth]{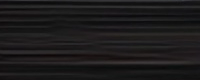} \hspace{-4.mm} &
\\ 
HR \hspace{-4.mm} &
LR \hspace{-4.mm} &
STAR+TLBVFI \hspace{-4.mm} &
STAR+BiM-VFI \hspace{-4.mm} &
\\
\includegraphics[width=0.190\textwidth]{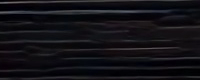} \hspace{-4.mm} &
\includegraphics[width=0.190\textwidth]{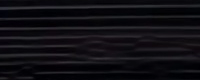} \hspace{-4.mm} &
\includegraphics[width=0.190\textwidth]{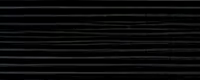} \hspace{-4.mm} &
\includegraphics[width=0.190\textwidth]{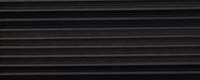} \hspace{-4.mm} &
\\ 
SeedVR+BiM-VFI \hspace{-4.mm} &
SeedVR2+BiM-VFI \hspace{-4.mm} &
VEnhancer \hspace{-4.mm} &
DiffST (ours) \hspace{-4mm}
\\
\end{tabular}
\end{adjustbox}
\\

\hspace{-0.48cm}
\begin{adjustbox}{valign=t}
\begin{tabular}{c}
\includegraphics[width=0.2215\textwidth]{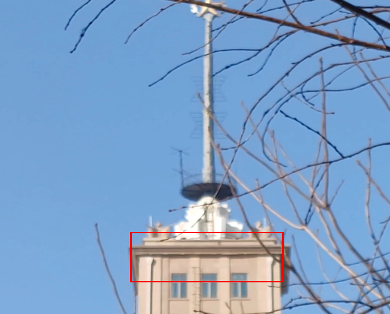}
\\
MVSR4x: 065
\end{tabular}
\end{adjustbox}
\hspace{-0.46cm}
\begin{adjustbox}{valign=t}
\begin{tabular}{cccccc}
\includegraphics[width=0.190\textwidth]{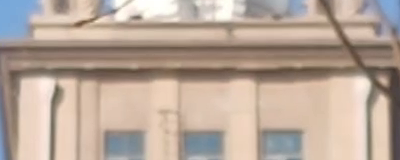} \hspace{-4.mm} &
\includegraphics[width=0.190\textwidth]{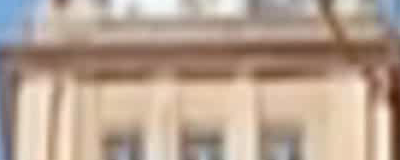} \hspace{-4.mm} &
\includegraphics[width=0.190\textwidth]{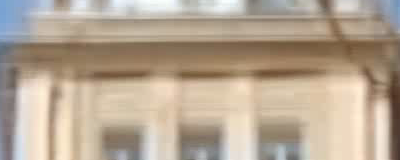} \hspace{-4.mm} &
\includegraphics[width=0.190\textwidth]{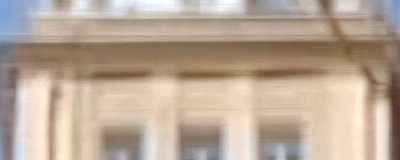} \hspace{-4.mm} &
\\ 
HR \hspace{-4.mm} &
LR \hspace{-4.mm} &
STAR+TLBVFI \hspace{-4.mm} &
STAR+BiM-VFI \hspace{-4.mm} &
\\
\includegraphics[width=0.190\textwidth]{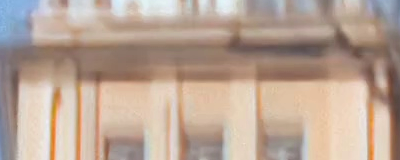} \hspace{-4.mm} &
\includegraphics[width=0.190\textwidth]{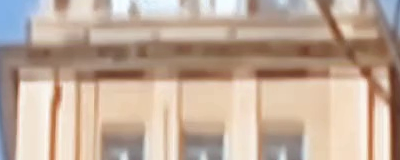} \hspace{-4.mm} &
\includegraphics[width=0.190\textwidth]{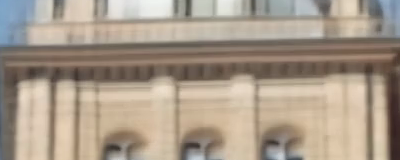} \hspace{-4.mm} &
\includegraphics[width=0.190\textwidth]{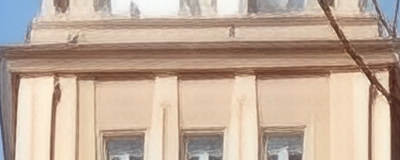} \hspace{-4.mm} &
\\ 
SeedVR+BiM-VFI \hspace{-4.mm} &
SeedVR2+BiM-VFI \hspace{-4.mm} &
VEnhancer \hspace{-4.mm} &
DiffST (ours) \hspace{-4mm}
\\
\end{tabular}
\end{adjustbox}
\\

\hspace{-0.48cm}
\begin{adjustbox}{valign=t}
\begin{tabular}{c}
\includegraphics[width=0.2215\textwidth]{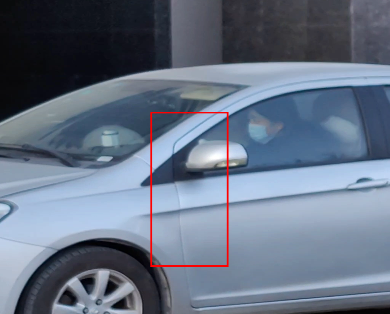}
\\
MVSR4x: 232
\end{tabular}
\end{adjustbox}
\hspace{-0.46cm}
\begin{adjustbox}{valign=t}
\begin{tabular}{cccccc}
\includegraphics[width=0.190\textwidth]{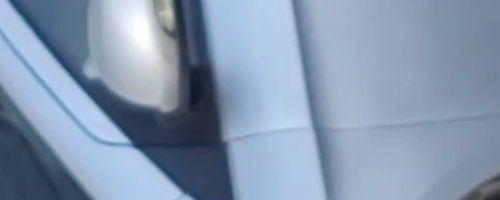} \hspace{-4.mm} &
\includegraphics[width=0.190\textwidth]{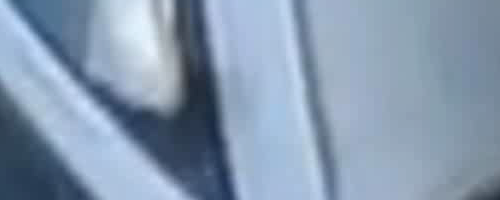} \hspace{-4.mm} &
\includegraphics[width=0.190\textwidth]{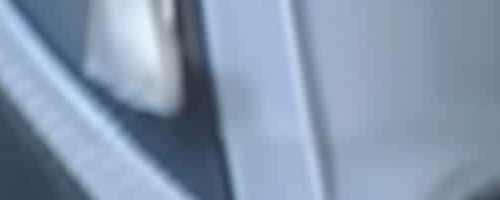} \hspace{-4.mm} &
\includegraphics[width=0.190\textwidth]{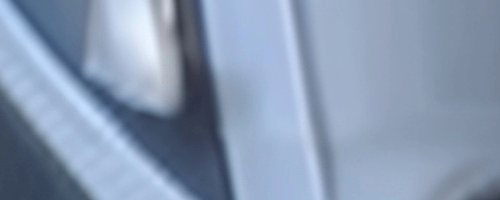} \hspace{-4.mm} &
\\ 
HR \hspace{-4.mm} &
LR \hspace{-4.mm} &
STAR+TLBVFI \hspace{-4.mm} &
STAR+BiM-VFI \hspace{-4.mm} &
\\
\includegraphics[width=0.190\textwidth]{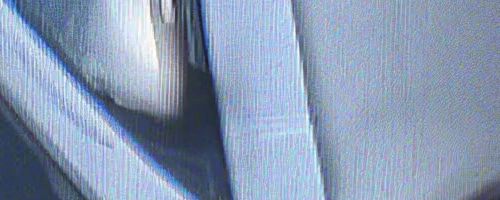} \hspace{-4.mm} &
\includegraphics[width=0.190\textwidth]{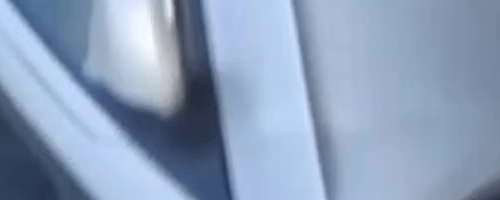} \hspace{-4.mm} &
\includegraphics[width=0.190\textwidth]{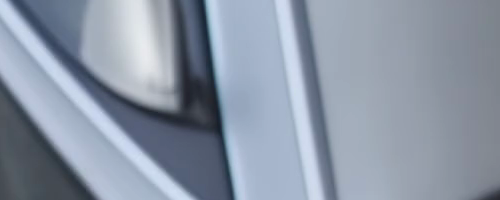} \hspace{-4.mm} &
\includegraphics[width=0.190\textwidth]{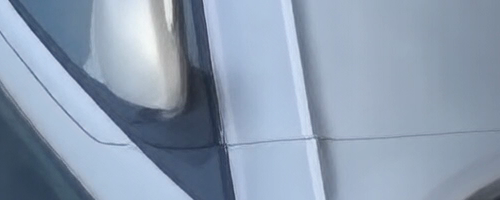} \hspace{-4.mm} &
\\ 
SeedVR+BiM-VFI \hspace{-4.mm} &
SeedVR2+BiM-VFI \hspace{-4.mm} &
VEnhancer \hspace{-4.mm} &
DiffST (ours) \hspace{-4mm}
\\
\end{tabular}
\end{adjustbox}

\end{tabular}
\caption{Visual comparison on real-world (MVSR4x~\cite{wang2023benchmark} and RealVSR~\cite{yang2021real}) datasets.}
\vspace{20.mm}
\label{fig:visual-supp-2}
\end{figure*}

We provide additional visual comparisons in Figs.~\ref{fig:visual-supp-1} and \ref{fig:visual-supp-2}. These examples extend the main qualitative comparison in Sec.~\ref{sec:sota} and Fig.~\ref{fig:visual}. Compared with existing methods, our approach reconstructs more realistic and accurate details. These results further demonstrate the effectiveness of our  DiffST.

\begin{table*}[t]
\scriptsize
\centering
\setlength{\tabcolsep}{.5mm}
\begin{tabular}{l|cccc|cccc|cccc} 
\toprule[0.15em]
\rowcolor{color3} &  \multicolumn{4}{c|}{Zooming Slow-Mo} & \multicolumn{4}{c|}{BF-STVSR} & \multicolumn{4}{c}{DiffST (ours)}\\
\rowcolor{color3} \multirow{-2}{*}{Datasets} & {PSNR $\uparrow$} & {LPIPS $\downarrow$} & {CLIP-IQA $\uparrow$} & {DOVER $\uparrow$} & {PSNR $\uparrow$} & {LPIPS $\downarrow$} & {CLIP-IQA $\uparrow$} & {DOVER $\uparrow$} & {PSNR $\uparrow$} & {LPIPS $\downarrow$} & {CLIP-IQA $\uparrow$} & {DOVER $\uparrow$} \\

\midrule[0.15em]
UDM10 & \textcolor{blue}{23.71} & \textcolor{blue}{0.5117} & \textcolor{blue}{0.1611} & 0.0863 & 23.65 & 0.5256 & 0.1601 & \textcolor{blue}{0.1011} & \textcolor{red}{24.92} & \textcolor{red}{0.2564} & \textcolor{red}{0.4086} & \textcolor{red}{0.7780} \\
Vid4 & \textcolor{red}{20.80} & \textcolor{blue}{0.5468} & 0.1908 & 0.1326 & 15.73 & 0.6221 & \textcolor{blue}{0.1925} & \textcolor{blue}{0.1343} & \textcolor{blue}{19.99} & \textcolor{red}{0.2699} & \textcolor{red}{0.2735} & \textcolor{red}{0.6076} \\
MVSR4x & 11.43 & 0.7037 & \textcolor{blue}{0.4540} & 0.0310 & \textcolor{red}{22.83} & \textcolor{blue}{0.4532} & 0.2467 & \textcolor{blue}{0.0985} & \textcolor{blue}{22.24} & \textcolor{red}{0.3320} & \textcolor{red}{0.4565} & \textcolor{red}{0.6739} \\
RealVSR & 9.71 & 0.8972 & 0.2227 & 0.0606 & \textcolor{red}{20.52} & \textcolor{blue}{0.3147} & \textcolor{blue}{0.2252} & \textcolor{blue}{0.4841} & \textcolor{blue}{19.01} & \textcolor{red}{0.2151} & \textcolor{red}{0.3833} & \textcolor{red}{0.8048} \\
\bottomrule[0.15em]
\end{tabular}
\caption{Quantitative comparison with more STVSR methods.}
\label{tab:more_method}
\end{table*}

\section{Explanations for Checklist}
\label{sec:checklist_explanations}

\subsection{Limitations}
In this work, we propose DiffST, an efficient one-step diffusion-based method for real-world space-time video super-resolution. While DiffST achieves strong performance and efficiency, it may still be limited under extremely severe degradations or complex motion. In addition, the VAE encoding and decoding process remains part of the overall computational cost.

\subsection{Broader Impacts}
DiffST improves the quality and efficiency of space-time video super-resolution. We do not foresee direct negative societal impacts from the proposed technical contributions.


\end{document}